\documentclass[11pt]{article}

\usepackage[T1]{fontenc}
\usepackage{lmodern}
\usepackage[utf8]{inputenc}
\usepackage{microtype}
\usepackage{framed}
\usepackage{listings}
\usepackage{vmargin}
\usepackage{setspace}
\usepackage{mathrsfs, mathenv}
\usepackage{amsmath, amsthm, amssymb, amsfonts, amscd}
\usepackage{graphicx}
\usepackage{epstopdf}
\usepackage[svgnames]{xcolor}
\usepackage{calc}
\usepackage{hyperref}
\hypersetup{citecolor=blue, colorlinks=true, linkcolor=black}
\usepackage[ruled, vlined]{algorithm2e}
\usepackage[capitalise]{cleveref}
\usepackage{subcaption}
\usepackage{bbm}
\usepackage{cite}
\usepackage{verbatim}
\usepackage{pgfplots}
\usepackage{tikz}
\usetikzlibrary{arrows,decorations.pathmorphing,backgrounds,positioning,fit,matrix}
\usepackage{etoolbox}
\usepackage{color}
\usepackage{lipsum}
\usepackage{ifthen}
\usepackage[title]{appendix}
\usepackage{autonum}
\usepackage{accents}
\usepackage{xpatch}
\usepackage[]{algpseudocode}
\usepackage{multicol}
\usepackage{multirow}
\usepackage[abs]{overpic}
\usepackage{eqnarray}
\usepackage{bm}

\usetikzlibrary{arrows.meta,calc,3d,positioning}

\crefname{assumption}{Assumption}{Assumptions}
\crefname{figure}{Figure}{Figures}
\theoremstyle{plain}

\numberwithin{equation}{section}

\theoremstyle{definition}

\newtheorem*{remark}{Remark}

\ifpdf
  \DeclareGraphicsExtensions{.eps,.pdf,.png,.jpg}
\else
  \DeclareGraphicsExtensions{.eps}
\fi

\usepackage{mathtools}
\usepackage{booktabs}

\usepackage{pgfplots}
\usepackage{tikz}
\usetikzlibrary{patterns,arrows,decorations.pathmorphing,backgrounds,positioning,fit,matrix}
\usepackage[labelfont=bf]{caption}
\usepackage{here}
\usepackage[font=normal]{subcaption}

\usepackage{tablefootnote}

\usepackage{enumitem}
\setlist[itemize]{leftmargin=.5in}
\setlist[enumerate]{leftmargin=.5in,topsep=3pt,itemsep=3pt,label=(\roman*)}

\newcommand{\TheTitle}{FalconBC:  Flow matching for Amortized inference of Latent-CONditioned physiologic Boundary Conditions}
\newcommand{\TheAuthors}{C. H. Choi, A. L. Marsden, D. E. Schiavazzi}
\title{\TheTitle}
\author{
Chloe H. Choi \thanks{Department of Mechanical Engineering, Stanford University, Stanford, CA. USA}
\and Alison L. Marsden \thanks{Institute for Computational and Mathematical Engineering, Stanford University, Stanford, CA, USA.} \thanks{Bioengineering, Stanford University, Stanford, CA, USA.} \thanks{Pediatric Cardiology, Stanford University, Stanford, CA, USA.} 
\and Daniele E. Schiavazzi \thanks{Department of Applied and Computational Mathematics and Statistics, University of Notre Dame, Notre Dame, IN, USA.}}
\date{}

\usepackage{amsopn}

\definecolor{shade}{RGB}{100, 100, 100}
\definecolor{bordeaux}{RGB}{128, 0, 50}

\usepackage[usestackEOL]{stackengine}

\definecolor{leg1}{RGB}{0,114,189}
\definecolor{leg2}{RGB}{217,83,25}
\definecolor{leg3}{RGB}{237,177,32}
\definecolor{leg4}{RGB}{126,47,142}
\definecolor{leg5}{RGB}{119,172,48}

\definecolor{leg21}{RGB}{62,38,169}
\definecolor{leg22}{RGB}{46,135,247}
\definecolor{leg23}{RGB}{55,200,151}
\definecolor{leg24}{RGB}{254,195,56}

\DeclareMathOperator*{\argmax}{arg\,max}

\ifpdf
\hypersetup{
	pdftitle={\TheTitle},
	pdfauthor={\TheAuthors}
}
\fi

\begin{document}
	
\maketitle

%\tableofcontents

\begin{abstract}

\noindent Boundary condition tuning is a fundamental step in patient-specific cardiovascular modeling. 
Despite an increase in offline training cost, recent methods in data-driven variational inference can efficiently estimate the joint posterior distribution of boundary conditions, with amortization of training efforts over clinical targets.
However, even the most modern approaches fall short in two important scenarios: open-loop models with known mean flow and assumed waveform shapes, and anatomies affected by vascular lesions where segmentation influences the reachability of pressure or flow split targets. In both cases, boundary conditions cannot be tuned in isolation.
We introduce a general amortized inference framework based on probabilistic flow that treats clinical targets, inflow features, and point cloud embeddings of patient-specific anatomies as either conditioning variables or quantities to be jointly estimated.
We demonstrate the approach on two patient-specific models: an aorto-iliac bifurcation with varying stenosis locations and severity, and a coronary arterial tree.

\end{abstract}


% ====================
\section{Introduction}
% ====================

Increased access to cost-efficient computational resources has facilitated the generation of high-fidelity, three-dimensional (3D) computational fluid dynamics (CFD) simulations that accurately capture patient-specific hemodynamics at the organ level~\cite{schwarz2023beyond}.
A fundamental contribution to this accuracy is the ability to select appropriate boundary conditions -- a problem we will refer to as \emph{boundary condition tuning} -- by using inexpensive reduced-order models, in an effort to approximate the impedance of downstream vascular networks beyond the anatomy of interest~\cite{vignon2006outflow}. Such models typically combine vascular resistance and compliance to reconcile the pulsatile nature of the aortic flow with steady distal flow, and can be as simple as a three-element Windkessel models~\cite{westerhof1971artificial}, i.e., formulated through one differential and two algebraic equations. In more complex scenarios, closed loop lumped parameter physiologic models are employed \cite{hu2025multiphysics,kung2014simulation}.

% the problem with tuning
% computational cost 
Tuning boundary conditions is a challenging problem for multiple reasons.
First, three-dimensional cardiovascular models are expensive, so inference with traditional sample-based estimators becomes quickly intractable, as they require sequential generations of thousands of model solutions.  As a result, deterministic optimization methods are often used. However, these approaches produce only point estimates and therefore fail to capture uncertainty and identifiability that are naturally part of posterior distributions. 
% re-tune due to new observations, new geometry or new inflow
Second, clinical data are often incomplete or inconsistent, necessitating parameter estimation in the presence of uncertainty and missing data. Datasets are typically noisy and may combine measurements of differing fidelity derived from multiple modalities, including echocardiography, magnetic resonance imaging (MRI), and catheterization.
% for open loop models, the coupling between bounbdary conditions, inflow or geometry
More interestingly, \emph{unknown} inflow waveform shape or \emph{inaccuracies} arising from noisy image acquisition or segmentation errors may affect the reachability of clinical targets, necessitating \emph{joint estimation} of boundary conditions, inflow and/or model anatomy. For example, uncertainties can arise from estimating cardiac output from echocardiograms or computed tomography (CT) data ~\cite{brault2017uncertainty,seresti2025validation} or from phase contrast MRI or 4DFlow MRI \cite{friman2011probabilistic,youssefi2018impact}. Previous work in \cite{MAHER21_geometric} also illustrated how segmentation uncertainty can influence simulation predictions.

% zero dimensional approximations of 3D cardiovascular models for tuning
To mitigate the computational cost of boundary condition tuning, inexpensive zero-dimensional approximations (0D, \emph{lumped parameter network}, or LPN models) can be used as a substitute of high-fidelity three-dimensional models. 
This is particularly appealing as LPNs can be automatically created from the three-dimensional model centerlines~\cite{piccinelli2009framework,pfaller22_centerline}, e.g., using SimVascular~\cite{simvascular}.
LPN models have been used to investigate congenital heart disease~\cite{kung2013predictive, migliavacca2006multiscale, hu2025multiphysics}, coronary artery disease \cite{kim2010developing, kim2010patient, grande20221d}, and aortic coarctation~\cite{nair2025experiments}, or as a replacement for 3D models in \emph{outer loop}, many-query tasks such as parameter optimization~\cite{brown2024modular, RUBIO2025hybrid, nair2024non, li2023fast, nair2023hemodynamics, ramazanli2025modeling}, multi-fidelity uncertainty quantification~\cite{schiavazzi2018multifidelity, seo2020multifidelity, seo2020effects, ZGS24, choi2025_aobif, FLEETER20_mlmf, richter2024bayesian, MZK24} or sensitivity analysis~\cite{schafer2024global}. However, these models are not always sufficiently accurate, as seen in \cite{pfaller22_centerline} which surveys the reconstruction accuracy of various pressure and flow quantities of interest, in \cite{nair2024non} which compares pressure drop across a stenosis in patient specific 3D and analogous 0D models of aortic coarctation, and in \cite{RUBIO2025hybrid} which demonstrates the use of machine learning-based strategies to improve 0D model accuracy.

% mitigate the computational cost of retraining: amortization
Recent advances in data-driven estimation of conditional distributions and amortized inference are making the retraining of boundary conditions increasingly cost-effective.
A particularly flexible tool we adopt in this paper is flow matching~\cite{lipman2024fm, tong2024cfm}, which can be derived as a continuous limit of normalizing flow~\cite{kobyzev2020normalizing, papamakarios2021normalizing}.
While traditional Markov Chain Monte Carlo (MCMC) methods~\cite{jackman2000estimation, haario1999adaptive, haario2006dram, gilks1994adaptive, storn1997differential, vrugt_2009, neal2011mcmc, vrugt_2012, hoffman2014no} require repeated evaluations of a simulator to draw posterior samples through a \emph{memoryless} process, \emph{conditional flow matching} (CFM) \emph{learns} to generate conditional distributions from joint samples and hence has the advantage of amortization.
Owing to its easy implementation and fast inference ability, CFM has gained interest in many applications, such as robotic manipulation policy learning~\cite{chisari2024learning}, segmentation of three-dimensional blood vessels~\cite{wittmann2025vesselfm}, and medical images~\cite{bogensperger2025flowsdf}.
While a large simulation dataset is still needed for offline training, fast generation of posterior samples can be performed online over multiple conditioning sets.
Note that this allows for joint estimation of boundary conditions and any arbitrary quantity that can be computed as an output of a physics-based model, and for which it is possible to generate a meaningful embedding.

In this study, we focus on two important aspects that often lead to lack of convergence during boundary condition tuning: the inclusion of Fourier features extracted from inflow waveforms, and the data-driven extraction of latent variables parameterizing changes in the lumen surface due to vascular lesions. 
Regarding the generation of a geometric embedding, existing methods register shapes as 3D point clouds~\cite{monji2023review}, as used in~\cite{liu2019flownet3d, croquet2021unsupervised, amor2022resnet}, and by means of deep implicit functions (e.g., continuous signed distance functions) expressed through neural networks~\cite{chen2019learning, xu2019disn, Park_2019_CVPR}, as used in~\cite{sun2022topology, KONG2024sdf4chd, lee2026accuracy}. 
The latter ideas have been used to register multiple geometries of the same topology based on a template shape in~\cite{croquet2021unsupervised, sun2022topology,KONG2024sdf4chd}, such as for a suite of healthy aorta models in~\cite{tenderini2025deform} and for biventricular geometries with tetralogy of Fallot in~\cite{martinez2025biv}.

% What we propose
In this study, we introduce \emph{FalconBC}, a CFM-based framework to predict distributions of boundary condition for cardiovascular models.
As an example case, we focus on an aorto-iliac bifurcation model with RCR-type BCs. 
We examine common boundary condition tuning scenarios from clinical targets, including systolic/diastolic pressures, flow split and mean branch flow. 
We investigate tuning of one total resistance and capacitance (two dimensions), to one RC for each of the two outlet (four dimensions fixing the proximal-to-distal resistance ratio), to one RCR circuit for each outlet (six dimensions).
We then generate inflow curves from a dataset and use the generated inflows to train the CFM model. 
Finally, we create a set of left and right iliac artery stenosis models, and encode the deformation from a point cloud representation of their lumen surface.
To the best of our knowledge, this is the first framework for amortized estimation of physiologic boundary conditions in cardiovascular modeling which can flexibly handle either conditioning on or jointly estimating inflow waveform features and point cloud embeddings of diseased anatomies.

This paper makes the following unique contributions:
\vspace{-5pt}
\begin{itemize}\itemsep 0pt
\item We propose a paradigm for amortized inference that does not require re-training for new clinical targets, inflow waveforms and lumen surface embedding, suggesting the possibility of a new \emph{foundational model} to streamline boundary condition tuning, which holds the potential to be incorporated for quick model adjustment in a digital twin scenario, or for boundary condition tuning over multiple anatomies.
\item We propose a data-driven encoder-decoder architecture to extract an embedding based on a point cloud characterization of the entire lumen surface, and demonstrate its use for both conditioning and joint estimation of boundary conditions.
\item We demonstrate a flow matching-based framework to solve \emph{generalized} boundary condition tuning problems, where additional relevant latent space features can be jointly estimated to improve reachability of clinical targets.
\end{itemize}
\vspace{-5pt}

% Section layout
The paper is organized as follows. In \cref{sec:methods}, we introduce CFM as applicable to boundary condition tuning and the encoder-decoder structure used to create interpretable latent spaces. In \cref{sec_examples}, we demonstrate the performance of CFM on a model of the aorto-iliac bifurcation, first considering an increasing number of estimated boundary conditions, then focusing on conditional and joint estimation of inflow features and lumen anatomy. We conclude with an example relevant to coronary artery disease. Finally, limitations and areas of future work are discussed in \cref{sec:conclusion}.

% ======================================
\section{Methodology}\label{sec:methods}
% ======================================

Consider the problem of generating samples from the distribution $q(\bm{x})$, where $\bm{x}\in\bm{\mathcal{X}}\subset \mathbb{R}^d$. 
Flow matching (FM~\cite{lipman2024fm}) is a recently proposed technique in generative modeling that achieves this by constructing a \emph{probability path}, $\rho_{t}(\bm{x}),\,t\in[0,1]$, from a known \emph{base} distribution $\rho_0(\bm{x}_0) = p(\bm{x})$ to the \emph{target} $\rho_1(\bm{x}_{1}) = q(\bm{x})$. 
To do so, one first approximates a velocity field $\bm{u}(\bm{x},t) = \bm{u}_{t}(\bm{x})$, $\bm{u}:\mathbb{R}^d \times [0,1] \rightarrow \mathbb{R}^d$ that \emph{generates} the probability path $\rho_{t}(\bm{x})$.
This means that the \emph{flow} $\bm{\psi}_{t} : \mathbb{R}^d \times [0,1] \rightarrow \mathbb{R}^d$ associated with the velocity $\bm{u}(\bm{x},t)$ is defined as a solution of the initial value problem
\begin{equation}\label{eq:fm_ode}
    \frac{\mathrm{d}}{\mathrm{d}t}\,\bm{\psi}_t(\bm{x}) = \bm{u}_t [\bm{\psi}_t(\bm{x})],\,\,\bm{\psi}_0(\bm{x})=\bm{x}_{0},\,\,\,\text{satisfies}\,\,\,\bm{x}_t \coloneq \bm{\psi}_t(\bm{x}_0) \sim \rho_t(\bm{x}),\,\,\text{for}\,\, \bm{x}_0 \sim p.
\end{equation}

In practice, we start with an easy-to-sample standard Gaussian base distribution $\bm{X}_0 \sim \mathcal{N}(\bm{0},\bm{I})$ and construct $\rho_t(\bm{x})$ using the law of total probability from the collection of conditional paths $\rho_{t|1}(\bm{x}|\bm{x}_1)$ using
\begin{equation}\label{eq:prob_path}
    \rho_t(\bm{x}) = \int_{\bm{\mathcal{X}}} \rho_{t|1}(\bm{x}|\bm{x}_1)\,q(\bm{x}_1)\,\mathrm{d}\bm{x}_1,\,\,\text{where}\,\,\rho_{t|1}(\bm{x}|\bm{x}_1) = \mathcal{N}(\bm{x}|t\,\bm{x}_1, (1-t^2)\,\bm{I}).
\end{equation}
Note that conditional samples $X_{t|1} = t\,\bm{x}_1 + (1-t)\bm{X}_0 \sim \rho_{t|1}$ consistent with~\eqref{eq:prob_path} induce a \emph{flow} with velocities
\begin{equation}\label{eq:rect_vels}
\bm{u}_{t|1}(\bm{x}|\bm{x}_1) = \frac{\bm{x}_{1} - \bm{x}}{1-t}.
\end{equation}
We now learn a parametric approximation $u^{\bm{\theta}}_t(\bm{x})$ for the velocity field $u_t(\bm{x})$ by minimizing a conditional flow matching loss of the form
\begin{equation}\label{eq:cfm_loss}
    \mathcal{L}_{\text{CFM}}(\bm{\theta}) = \mathbb{E}_{t,\bm{X}_t,\bm{X}_1} \| \bm{u}_t^{\bm{\theta}}(\bm{X}_t) - \bm{u}_{t|1}(\bm{X}_{t}|\bm{X}_{1}) \|^2,\,\,\text{where}\,\, t \sim \mathcal{U}[0,1],\,\,\text{and}\,\, X_t \sim \rho_t.
\end{equation}
Specifically, we use multilayer perceptrons (MLPs, see, e.g.,~\cite[Chapter 6]{DeepLearning}), determining the optimal values of their weights and biases through gradient descent using Adam~\cite{adam_opt_2017}. We include details on the bounds of hyperparameter optimization, which we perform using Optuna~\cite{Optuna}, in~\cref{app:optuna_hyperparams}.

Sampling from conditional distributions can be achieved with minimal additional effort.
Suppose an additional set of variables $\bm{Y} \in \mathbb{R}^m$ is available and comprised of, e.g., clinical targets, inflow features, or an embedded representation of a patient-specific anatomy.
We can then estimate \emph{conditional velocities} by amortization of the parameters $\bm{\theta}$ over $\bm{Y}$, minimizing a loss of the form
\begin{equation}\label{eq:cfm_loss}
    \mathcal{L}_{\text{CFM}}(\bm{\theta}) = \mathbb{E}_{t,\bm{X}_t,\bm{X}_1,\bm{Y}} \| \bm{u}_t^{\bm{\theta}}(\bm{X}_t|\bm{Y}) - \bm{u}_{t|1}(\bm{X}_{t}|\bm{X}_{1},\bm{Y}) \|^2.
\end{equation}
The accuracy of the conditional distributions learned through flow matching is evaluated as follows. First we generate a solution dataset combining inputs $\bm{x}\in\bm{\mathcal{X}}$ and conditional features $\bm{y}\in\bm{\mathcal{Y}}$, resulting in $(\bm{x}^{(j)},\bm{y}^{(j)})_{j=1}^{N}$.
We then train CFM using the \emph{training dataset} $(\bm{x}^{(j)},\bm{y}^{(j)})_{j=1}^{N_{\text{train}}}$.
We then use the conditional features in the \emph{testing dataset} $\bm{y}^{(j)},\,j=N_{\text{train}}+1,\dots,N_{\text{train}}+N_{\text{test}}$ to generate distributions of input parameters $\widehat{\bm{x}}^{(k)},\,k=1,\dots,N_{\text{s}}$, ($N_s$ samples for each conditional feature in the testing dataset) employing rejection sampling to ensure satisfaction of known physical constraints (i.e., prior bounds).
Finally, samples $\widehat{\bm{y}}^{(k)},\,k=1,\dots,N_{s}$ are generated from a \emph{posterior predictive distribution}, by solving a cardiovascular model with boundary conditions from $\widehat{\bm{x}}^{(k)}$.

To assess successful amortization, we compute a mean absolute reconstruction error as:
\begin{equation}\label{equ:eps}
\begin{split}
\varepsilon_{i,\text{abs}}^{(j)} &= \frac{1}{N_s} \sum_{k=1}^{N_s} \left| \widehat{y}_i^{(k)} - y_i^{(j)} \right|,\,\,
    \overline{\varepsilon}_{i,\text{abs}} = \frac{1}{N_{\text{test}}} \sum_{j=1}^{N_{\text{test}}} \varepsilon_{i,\text{abs}}^{(j)}\\
\varepsilon_{i,\text{sign}}^{(j)} &= \frac{1}{N_s} \sum_{k=1}^{N_s} \left( \widehat{y}_i^{(k)} - y_i^{(j)} \right),\,\,
    \overline{\varepsilon}_{i,\text{sign}} = \frac{1}{N_{\text{test}}} \sum_{j=1}^{N_{\text{test}}} \varepsilon_{i,\text{sign}}^{(j)},  
\end{split}    
\end{equation}
where $N_\text{test}$ is set as 10\% of the total dataset size, $N_{s}$ is the number of posterior samples generated from CFM, $y^{(j)}_{i}$ is the $i$-th conditional feature in the true underlying $j$-th testing sample, and $\widehat{y}^{(k)}_{i}$ is the $i$-th output component from a cardiovascular model with boundary conditions $\widehat{\bm{x}}^{(k)}$. 
Relative errors are defined by dividing $(\widehat{y}_i^{(k)} - y_i^{(j)})$ or $|\widehat{y}_i^{(k)} - y_i^{(j)}|$ by $y_i^{(j)}$ or $|y_i^{(j)}|$, respectively.
Note that \eqref{equ:eps} combines the posterior predictive samples $\widehat{y}_i^{(k)},\,k=1,\dots,N_{s}$ with the true validation target $y_i^{(j)},j=1,\dots,N_{\text{test}}$. Consequently, while $\varepsilon_{i,\text{sign}}^{(j)}$ is expected to become increasingly small as $N$ increases, the same does not hold for $\varepsilon_{i,\text{abs}}^{(j)}$.

The model is trained via conditional flow matching to learn the flow between a base distribution and the posterior, without performing explicit likelihood evaluation.
Posterior samples are obtained by forward integration of the neural ODE \eqref{eq:fm_ode} using the adaptive Dormand–Prince 8-7 solver from \texttt{torchdiffeq}~\cite{chen2018neural}, and marginal densities are estimated empirically from the generated samples using histogram-based density estimation. 
Since our primary goal is to characterize the posterior distributions of the parameters, forward integration is sufficient to obtain representative samples; backward integration for likelihood evaluation is not necessary in this study, though it is feasible within the same framework if explicit log-densities are required (see, e.g, Sec. 3.6 in~\cite{lipman2024fm}).

% ===================
\subsection{FalconBC}
% ===================

\begin{figure}[!htb]
    \centering
    \includegraphics[width=\linewidth]{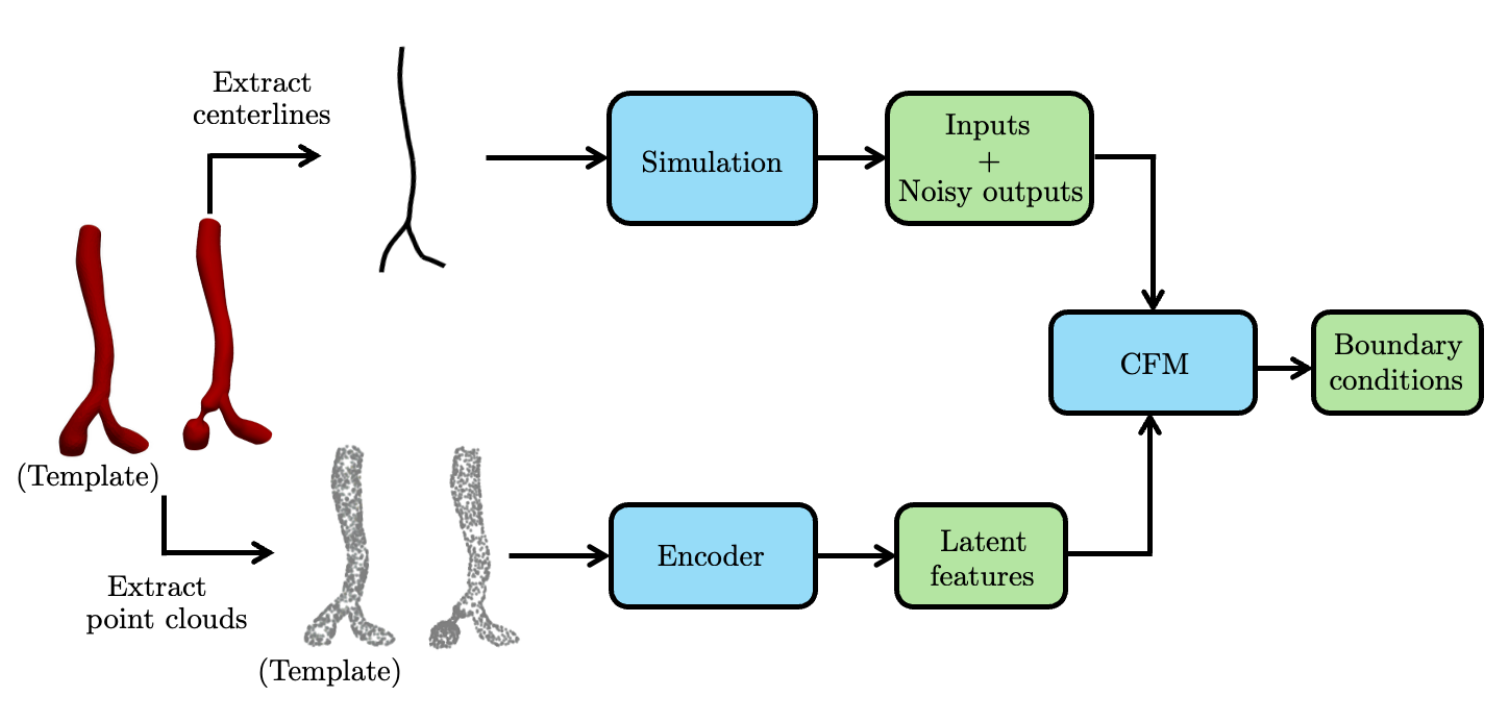}
    \caption{FalconBC workflow.}
    \label{fig:workflow}
\end{figure}

We illustrate the overall workflow for FalconBC in \cref{fig:workflow} and summarize its main steps in \cref{alg:falconbc}. 

\begin{algorithm}
\caption{Amortized inference of boundary conditions in FalconBC.}\label{alg:falconbc}
\textbf{Input:} Dataset $\{\mathcal{B},\mathcal{C},\mathcal{Z}\}$ (see Section~\ref{sec:cfm}) where $\mathcal{B}$ contains boundary condition samples, $\mathcal{C}$ contains noisy clinical targets or inflow features, and $\mathcal{Z}$ contains samples from the anatomical embedding introduced in Section~\ref{sec:geo_embedding}.\\
\textbf{Inference task setup:} The dataset $\{\mathcal{B},\mathcal{C},\mathcal{Z}\}$ is partitioned as $\{\bm{X},\bm{Y}\}$ based on which quantities are provided in the conditioning set ($\bm{Y}$), and which quantities need to be estimated ($\bm{X}$).\\
\textbf{Training FalconBC:} CFM is trained from $\{\bm{X},\bm{Y}\}$ by minimizing~\ref{eq:cfm_loss}.\\
\textbf{Evaluation:} For each set $\bm{Y}$, samples are generated from the posterior distribution of $\bm{X}$ from the numerical solution of \eqref{eq:fm_ode}. Amortization implies that the same CFM instance (no re-training required) will be used to generate the posterior for $\bm{X}$, under a varying conditioning set $\bm{Y}$.
\end{algorithm}

% ================================================
\subsubsection{Iliac artery stenosis model cohort}
% ================================================

Throughout this paper, we use a healthy aorto-iliac bifurcation model from~\cite{choi2025_aobif}. 
We first validate the posterior distribution from the paper in \cref{sec:validation}. 
We then use this geometry to investigate the effect of varying the boundary conditions from two parameters (i.e., total resistance and total capacitance) to six parameters (i.e., proximal/distal resistance and capacitance per branch) and training data set size in \cref{sec:bifurcation_BCs,sec:RCR_bc_inflow,,sec:4_BC_params}.

In addition, \cref{sec:RCR_bc_inflow} discusses a problem that has received limited attention in prior literature, i.e., joint estimation of boundary conditions and inflow waveform features. 
This is particularly valuable in two scenarios: when a waveform has been acquired with inherent uncertainty (e.g., from phase-contrast or 4D Flow MRI), or when a bulk flow measure such as cardiac output is available from another modality (e.g., echocardiography) but no waveform has been recorded. Importantly, this approach offers a significant improvement over the common practice of simply scaling an existing waveform to match its known statistics, such as cardiac output.

\begin{figure}[!htb]
    \centering

    % ---------- Left column (VERTICALLY CENTERED) ----------
    \begin{minipage}[c]{0.27\textwidth}
        \centering
        \begin{subfigure}{\linewidth}
            \centering
            \includegraphics[width=\linewidth]{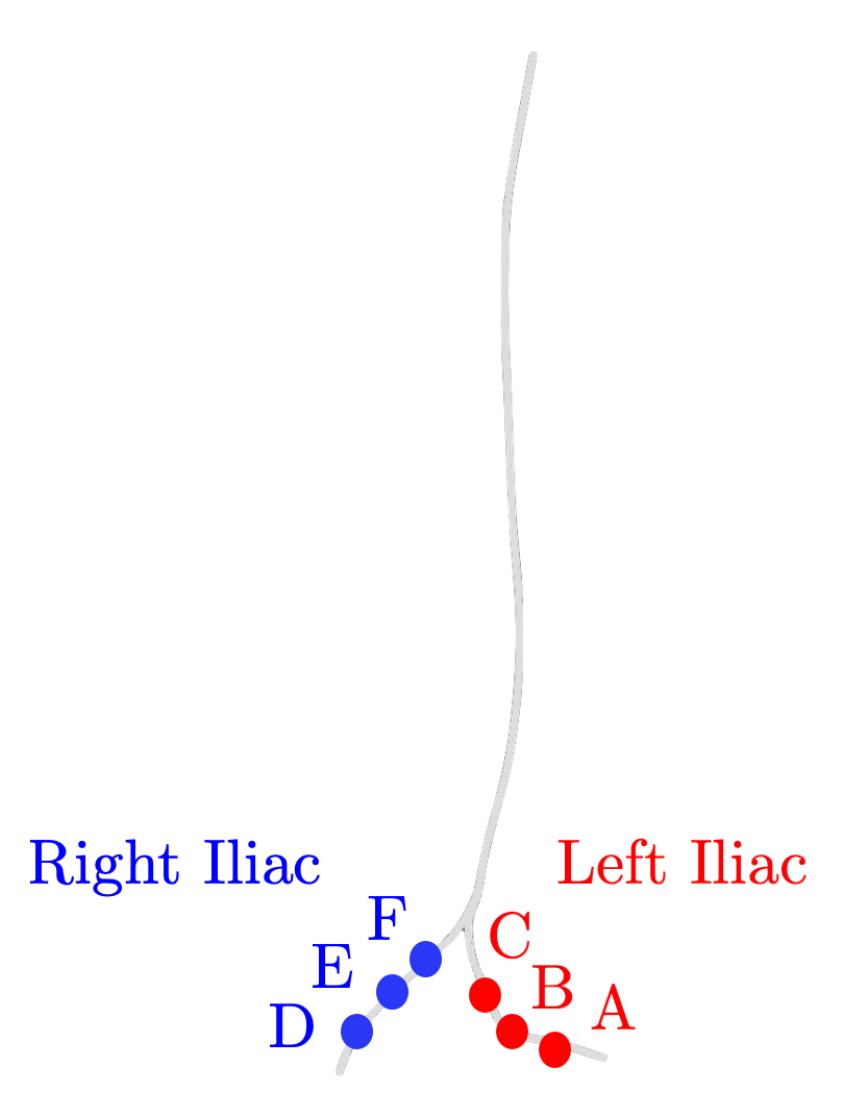}
            \caption{Stenosis locations at left and right iliac branches.}
            \label{fig:subfig_stenosis_locations}
        \end{subfigure}
    \end{minipage}
    \hfill
    % ---------- Right column (stacked, top-aligned) ----------
    \begin{minipage}[t]{0.7\textwidth}
        \centering

        \begin{subfigure}{\linewidth}
            \centering
            \includegraphics[width=\linewidth]{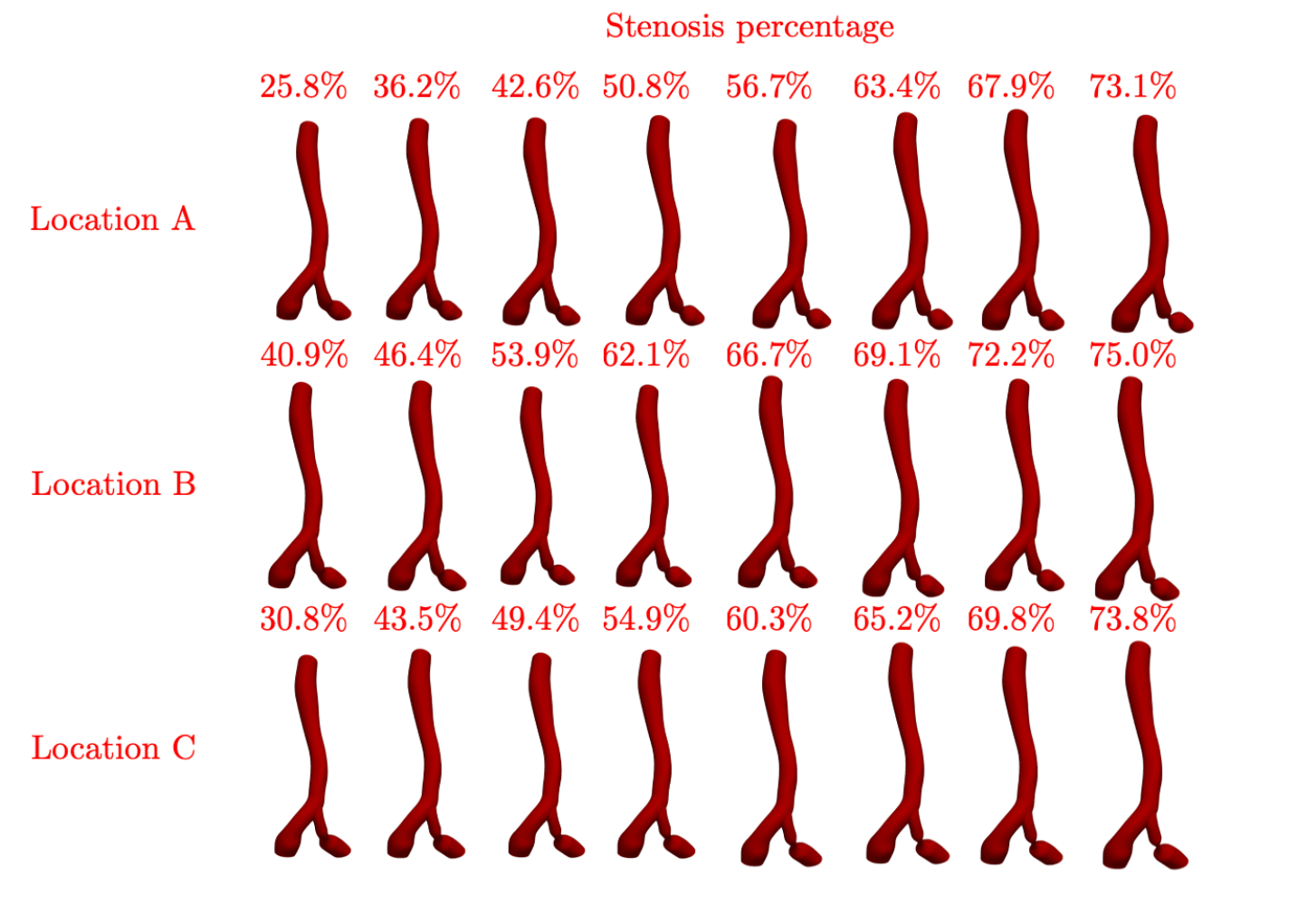}
            \caption{Diseased anatomies with varying degrees of left iliac artery stenosis.}
            \label{fig:subfig_left_stenoses}
        \end{subfigure}

        \vspace{0.5em}

        \begin{subfigure}{\linewidth}
            \centering
            \includegraphics[width=\linewidth]{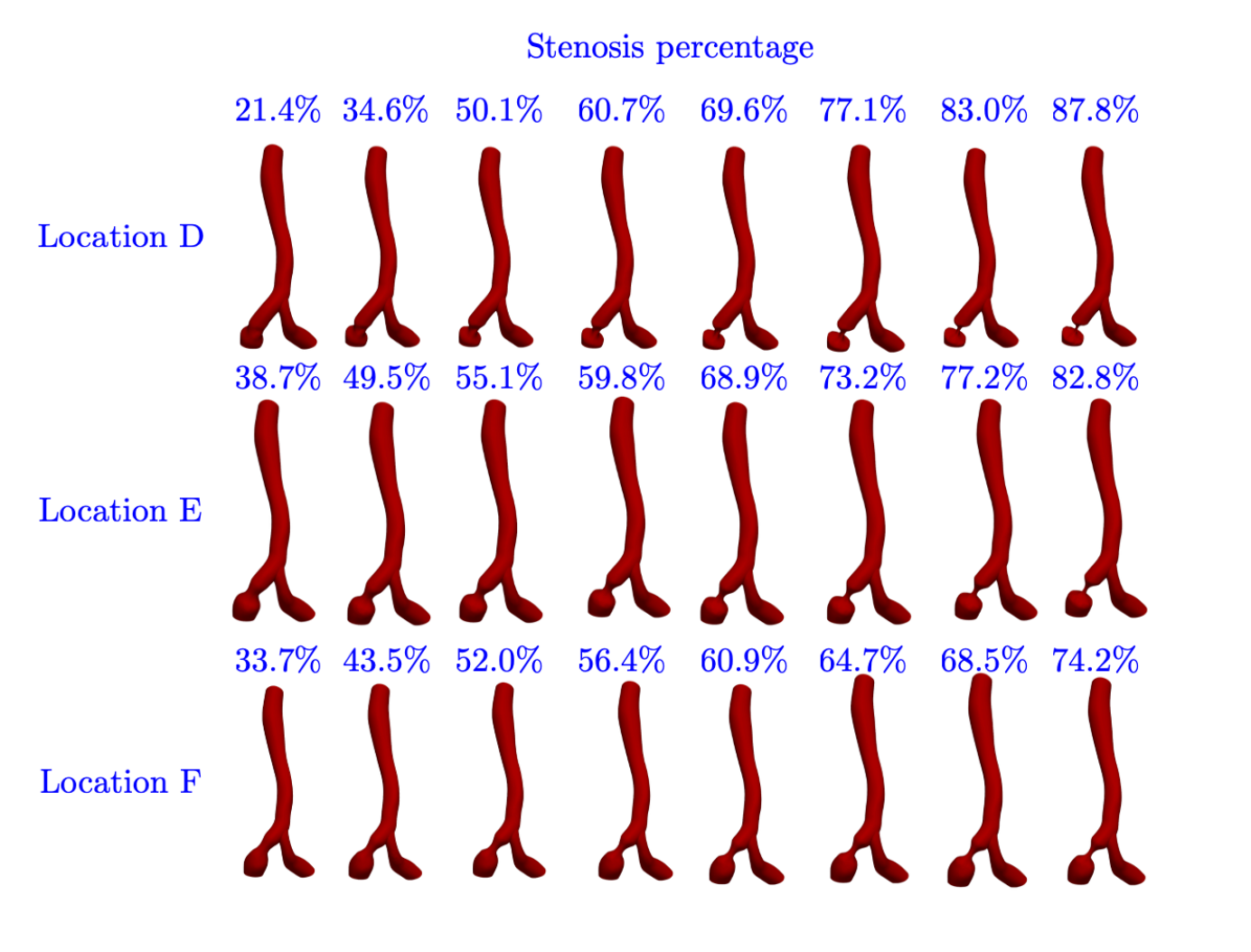}
            \caption{Diseased anatomies with varying degrees of right iliac artery stenosis.}
            \label{fig:subfig_right_stenoses}
        \end{subfigure}

    \end{minipage}

    \caption{Collection of diseased models with left and right iliac artery stenoses.}
    \label{fig:stenosis_suite}
\end{figure}

Finally, we consider the even more challenging problem of conditioning on and estimating anatomical features, as characterized by embedded representations derived from a point cloud characterization of the lumen surface.
Leveraging a recent update of \texttt{svMorph}~\cite{pham2023svmorph}, we construct $N_{\text{geom}}=48$ different geometries where a stenotic lesion of increasing severity is sculpted at three different locations along the left and right iliac arteries, as shown in \cref{fig:stenosis_suite}. 
We focus on left and right iliac artery stenosis, as these sites represent some of the most common locations of peripheral artery disease in affected patients~\cite{aaa_iliac_stenosis_19}.

\begin{figure}[!htb]
    \centering
    % Row 1: Two subfigures side-by-side
    \begin{subfigure}[t]{0.475\textwidth}
        \centering
        \includegraphics[width=\linewidth]{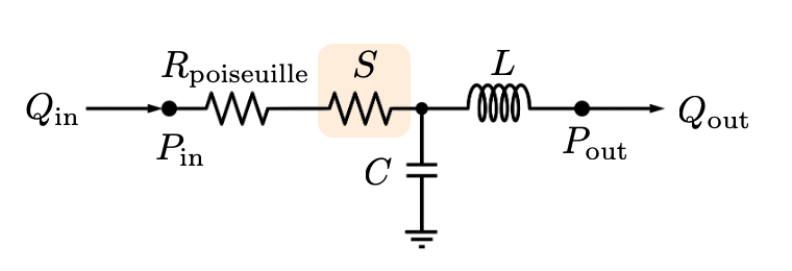}
        \caption{Zero-dimensional model of a generic blood vessel with stenosis in orange.}
        \label{fig:subfig_blood_vessel}
    \end{subfigure}
    \hfill
    \begin{subfigure}[t]{0.475\textwidth}
        \centering
        \includegraphics[width=0.9\linewidth]{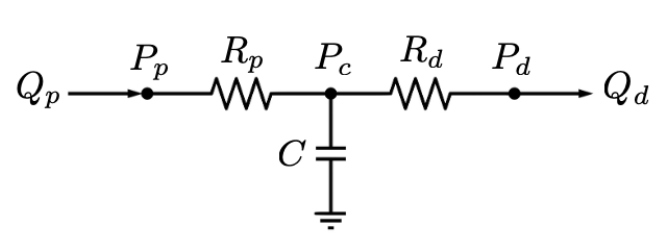}
        \caption{RCR outlet boundary condition.}
        \label{fig:subfig_rcr}
    \end{subfigure}
    \caption{Zero-dimensional model components for the selected aorto-iliac bifurcation model.}
    \label{fig:aobif_0d}
\end{figure}

Zero-dimensional (0D) models are generated in \texttt{SimVascular}~\cite{simvascular} from the stenotic geometries. As part of the 0D model construction workflow, the stenosis detection tool described in~\cite{pfaller22_centerline} is applied to automatically identify lesion locations, and incorporate empirically derived pressure–flow corrections at the stenoses.
Circuit-based approximations derived from zero-dimensional models are shown in \cref{fig:aobif_0d}, and consist of resistors, capacitors, inductors, and nonlinear resistors used to capture the relation between flow rate and pressure drop across a stenosis. Outlet boundary conditions are implemented as RCR circuits.

Nominal values of the outlet boundary conditions are obtained by keeping the blood vessel circuit parameters constant and performing optimization and inference using Nelder-Mead in \cref{sec:bifurcation_BCs,sec:RCR_bc_inflow} and differential evolution in \cref{sec:geometry_conditioning}, respectively. 
Details of the former are included in \cref{app:geometry_bc_optimization}. 
We then uniformly sample each of the boundary conditions around their nominal values independently, with perturbations equal to $\pm$30\%. Using these boundary conditions, we run zero-dimensional simulations to assemble a dataset of systolic and diastolic inlet pressures and the mean outlet flows.
To account for irreducible aleatoric measurement noise, we add a heteroskedastic noise with distribution $\mathcal{N}(0, \boldsymbol{\Sigma})$, where $\boldsymbol{\Sigma}=\text{diag}( (0.05\,P_{\text{dia}})^2, (0.05\,P_{\text{sys}})^2, (0.1\, Q_{\text{mean,left}})^2, (0.1\, Q_{\text{mean,right}})^2)$. 
These noisy outputs are added to the conditioning set in FalconBC to infer the conditional distribution of the boundary conditions that, when applied to the model outlet, would produce such pressures and flows as outputs. 

% =================================================
\subsubsection{Creation of anatomical point clouds}
% =================================================

Each vascular geometry is provided as a tetrahedral mesh. We extract vertex coordinates from the surface mesh, $\mathcal{M}$, and form a point cloud $\bm{\mathcal{S}} = \{ \bm{s}^{(n)} \}_{n=1}^{N_{p}} \subset \mathbb{R}^3$. 
In particular, consider the \emph{template} or baseline anatomy (typically a healthy anatomy without vascular lesions) as $\bm{\mathcal{T}} = \{\bm{t}^{(n)}\}_{n=1}^{N_{t}} \subset \mathbb{R}^3$.
We normalize each anatomy with respect to the template using
\begin{equation}
    \widetilde{\bm{s}}^{(n)} = \frac{\bm{s}^{(n)} - \bm{c}_{\bm{\mathcal{T}}}}{\bm{s}_{\bm{\mathcal{T}}}}, \quad n = 1,\dots,N_{p},
\end{equation}
where $\bm{c}_{\bm{\mathcal{T}}}$ and $\bm{s}_{\bm{\mathcal{T}}}$ denote the geometric center and scaling factor of the template, defined as
\begin{equation}
    \bm{c}_{\bm{\mathcal{T}}} = \dfrac{\min_n \bm{t}^{(n)} + \max_n \bm{t}^{(n)}}{2}, \,\,
    \bm{s}_{\bm{\mathcal{T}}} = \max\big(\max_n \bm{t}^{(n)} - \min_n \bm{t}^{(n)}\big)/2.
\end{equation}
This transformation ensures that all geometries are centered at the origin and lie within the cube
\([-1,1]^3\), providing a consistent registration and scaling for downstream processing.
Subsequently, we uniformly sample a point cloud of size $N=1024$ and augment the training set by generating 50 such point clouds per geometry.
Doing so increases robustness in the representation of each geometry, with the ability to handle diversity due to the model generation and meshing process.

% ==========================================================
\subsubsection{Embedded representations of diseased vascular anatomies}\label{sec:geo_embedding}
% ==========================================================

Two different methods are discussed in this section for generating anatomical embeddings from lumen surface point clouds for anatomies with stenotic lesions.
First, we introduce a natural parameterization defined through an ordered set of locations of the form
\begin{equation}
\bm{z} = \big[ \text{left}_A,\, \text{left}_B,\, \text{left}_C,\, \text{right}_D,\, \text{right}_E,\, \text{right}_F \big]^{T} \in \mathbb{R}^6.
\end{equation}
Each component in $\bm{z}$ corresponds to a stenosis severity percentage, and only one component of $\bm{z}$ is non-zero at each realization. 

A second parameterization is constructed through a permutation-invariant PointNet-based~\cite{qi2017pointnet} encoder to \emph{learn} such an embedding.
We associate each input point cloud $\bm{\mathcal{S}}$ with a \emph{mode} $\bm{m} \in \{\bm{e_{i}} \in \mathbb{R}^6|i=1,\dots,6\},\,$  (anatomical location corresponding to $A$ through $E$ in Figure~\ref{fig:stenosis_suite}) where $\bm{e_{i}}$ is the $i$-th canonical basis vector in $\mathbb{R}^{6}$ and scalar \emph{severity} $\phi \in [0,1]$.

We employ a \emph{permutation-invariant} encoder $E_\theta = (E_\theta^{\text{mode}}, E_\theta^{\text{sev}})$ to extract global features from a point cloud $\bm{\mathcal{S}} = \{\bm{s}^{(n)}\}_{n=1}^{N}$, combining two steps
\begin{equation}
\widehat{\bm{m}} = E_\theta^{\text{mode}}(\bm{\mathcal{S}}) \in \mathbb{R}^6,\,\,\text{and}\,\,
\widehat{\psi} = \sigma\big(E_\theta^{\text{sev}}(\bm{\mathcal{S}})\big) \in [0,1], %\,\,\bm{s}\in\bm{\mathcal{S}},
\end{equation}
where $\mathbf{m}$ are the raw mode logits whose softmax defines a predicted probability distribution over the anatomical modes, $\sigma(\cdot)$ denotes a sigmoid function, and $\widehat{\psi}$ represents the predicted stenosis severity. 
The latent vector $\bm{z}$ is then constructed by combining the predicted severity with the softmax of the mode logits in training and the one-hot embedding in offline evaluation, as
\begin{equation}
\widehat{\bm{z}} = \begin{cases}
    \widehat{\psi} \cdot \mathrm{softmax}(\widehat{\bm m}) & \text{ training}, \\
    \widehat{\psi} \cdot \mathrm{oneHot}(\argmax_i (\mathrm{softmax}(\widehat{\bm m}))) & \text{ evaluation.}
\end{cases}
\end{equation}

Each template point $\bm{t}^{(n)} \in [-1,1]^3,\,n=1,\dots,N_{t}$ is encoded using $F$ high-dimensional Fourier features
\begin{equation}
\gamma(\bm{t}^{(n)}) = \big[ \sin(\omega_\ell\,\bm{t}^{(n)}), \cos(\omega_\ell\,\bm{t}^{(n)}) \big]_{\ell=0}^{F-1} \in \mathbb{R}^{6\cdot F},\,\,\omega_\ell = 2^\ell \pi,\,\, \ell = 0,\dots,F-1.
\end{equation}

\begin{figure}[!htb]
\centering
\includegraphics[width=\linewidth]{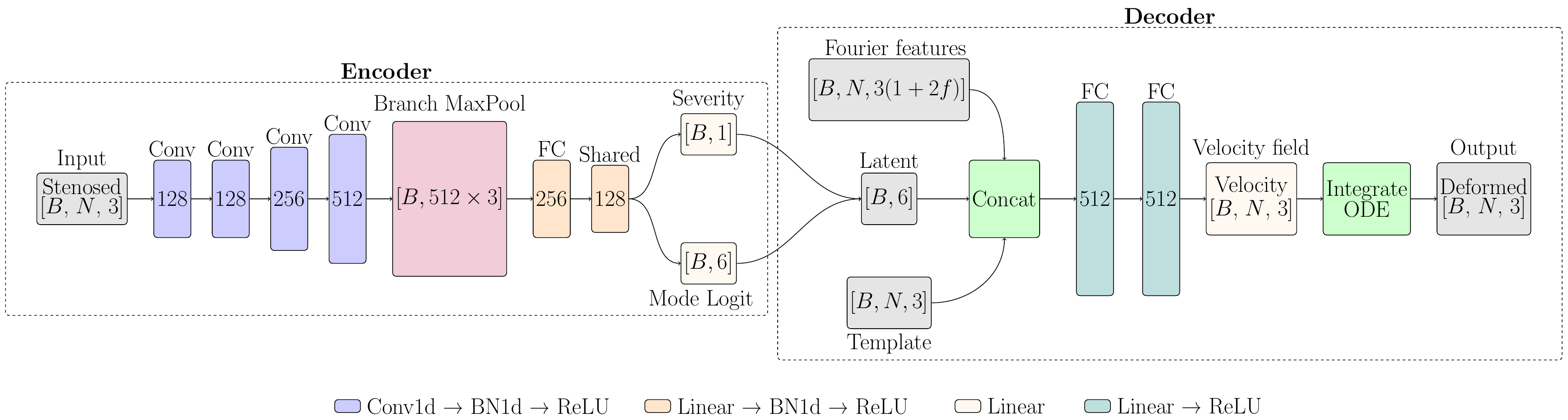}
\caption{\textbf{Encoder/decoder diagram}. Encoder produces latent vector $\bm{z}$; template points (below $\bm{z}$) and Fourier features (above $\bm{z}$) feed into decoder via a concatenation block. Encoder and decoder are highlighted with dashed boxes.}
\label{fig:encoderdecoder}
\end{figure}

% The decoder $D_\phi$ is a pointwise MLP mapping Fourier features, template coordinates, and latent vector encoding stenosis location and severity to \emph{diseased} point coordinates:
% \begin{equation}
% \widehat{\bm{s}}^{(n)} = D_\phi\big( \gamma(\bm{t}^{(n)}), \bm{t}^{(n)}, \bm{z} \big),\,\, n = 1,\dots,N,\,\,\text{and}\,\,\widehat{\bm{\mathcal{S}}} = \{\widehat{\bm{s}}^{(n)}\}_{n=1}^N,
% \end{equation}
% %
The decoder $D_\phi$ is a pointwise MLP that maps template points to deformed coordinates by internally predicting and integrating a velocity field, conditioned on Fourier features and a latent vector encoding stenosis location and severity. Starting from the template points, this integration is performed over $J$ steps:

\begin{equation}
\bm{s}^{(n)}_{j+1} = \bm{s}^{(n)}_j + h\, D_\phi\big(\gamma(\bm{s}^{(n)}_j), \bm{s}^{(n)}_j, \widehat{\bm{z}}\big), \quad j=0,\dots,J-1, \quad n = 1,\dots,N,
\end{equation}

where $\bm{s}^{(n)}_0 = \bm{t}^{(n)}$ are the template points, $\gamma(\cdot)$ are Fourier features, and $h = 1/J$ is the integration step. This simple first-order method has been shown to be effective in capturing deformations in \cite{tenderini2025deform,KONG2024sdf4chd}. The final deformed shape is
\[
\widehat{\bm{\mathcal{S}}} = \{\bm{s}^{(n)}_J\}_{n=1}^N.
\]
The system is trained end-to-end using a loss function of the form
\begin{equation}
\mathcal{L} = \mathcal{L}_{\mathrm{CD}} + \lambda_{\mathrm{sev}}\,\mathcal{L}_{\mathrm{sev}} + \lambda_{\mathrm{mode}}\,\mathcal{L}_{\mathrm{mode}},
\end{equation}
where $\lambda_{\mathrm{sev}}, \lambda_{\mathrm{mode}}$ are penalty hyperparameters.
The three components in the loss function represent the distance between the true and reconstructed diseased anatomies, as quantified by the \emph{symmetric Chamfer distance}~\cite{barrow1977parametric}
\begin{equation}
\mathcal{L}_{\mathrm{CD}}(\widehat{\bm{\mathcal S}}, \bm{\mathcal S})
=
\frac{1}{|\widehat{\bm{\mathcal S}}|}
\sum_{\widehat{\bm s} \in \widehat{\bm{\mathcal S}}}
\min_{\bm s \in \bm{\mathcal S}}
\|\widehat{\bm s} - \bm s\|_2
+
\frac{1}{|\bm{\mathcal S}|}
\sum_{\bm s \in \bm{\mathcal S}}
\min_{\widehat{\bm s} \in \widehat{\bm{\mathcal S}}}
\|\bm s - \widehat{\bm s}\|_2,
\end{equation}
the mean squared error on the predicted stenosis severity 
\begin{equation}
\mathcal{L}_{\mathrm{sev}} = (\widehat{\psi} - \psi)^2,
\end{equation}
and a cross-entropy loss on the stenosis location
\begin{equation}
\mathcal{L}_{\mathrm{mode}} = \mathrm{CE}(\widehat{\bm{m}}, \bm{m}),
\end{equation}
where $\psi$ and $\bm{m}$ are the true underlying degree of stenosis and mode, respectively. 
The final predicted shape is
\begin{equation}
\widehat{\bm{\mathcal{S}}} = D_\phi(\bm{\mathcal{T}}, \bm{z}),\,\,\bm{z} = \begin{cases}
    \psi \cdot \mathrm{softmax}(E_\theta^{\text{mode}}(\bm{\mathcal{S}})) & \text{ training}, \\
    \psi \cdot \mathrm{oneHot}(\argmax_i (\mathrm{softmax} ( E_\theta^{\text{mode}}(\bm{\mathcal{S}})))) &  \text{ evaluation.}
\end{cases}
%\psi\cdot\mathrm{oneHot}(E_\thetas{\text{mode}}(\bm{\mathcal{S}})),
\end{equation}
where $\bm{\mathcal{T}}$ is the template point cloud.
The network is trained jointly on anatomy, lesion severity and location, enabling \emph{template-conditioned, mode-specific deformations}.
The overall encoder-decoder framework is shown in \cref{fig:encoderdecoder}. The encoder receives only the 3D point cloud and a branch partition label per point. 
A PointNet-style backbone extracts permutation-invariant local features and branch-wise pooling isolates anatomical regions, ensuring that deformations remain spatially localized in feature space. The shared $1\times1$ convolutions implement a permutation-equivariant pointwise feature extractor, while the branch-wise max pooling provides permutation invariance within each anatomical region. The encoder then predicts (i) a categorical branch identity via cross-entropy supervision and (ii) a continuous severity via regression. 
Importantly, a single encoder is trained to learn the latent representation from the diseased surface geometry. At inference time, embeddings for unseen anatomies are obtained through a single forward pass. 
The latent representation is explicitly structured as a one-hot 6D vector, where only the active anatomical mode is scaled by the predicted severity. This architectural constraint removes rotational ambiguity and enforces disentanglement between stenosis location and magnitude. Thus, each stenosis location corresponds to a one-dimensional subspace in the latent space, reflecting the single intrinsic degree of freedom (severity) per stenosis. The decoder then maps this structured latent into a deformation of a template, effectively learning a low-dimensional anatomical deformation basis. This construction enforces disentanglement between location and severity and allows interpolation to unseen stenoses.

In this work, we used a one-hot vector to demonstrate the use of interpretable latent vectors. Note that, to flexibly encode point clouds of different patients with the same topology, one could use learned embeddings rather than an encoder and replace the decoder with that introduced in \cite{tenderini2025deform}. To encode point clouds of different patients with different topology, one could use the same encoder, use a more flexible latent space parameterization, and incorporate a template-free decoder (e.g., \cite{yang2019pointflow,peng2021shape}), where disentanglement would need to be carefully considered.
%The decoder is included solely to enforce geometric consistency in the learned latent space via point-set reconstruction. Explicit surface or mesh reconstruction is outside the scope of this work.

% ==========================================
\subsubsection{CFM framework}\label{sec:cfm}
% ==========================================

After training the encoder-decoder framework, the latent vector $\bm{z}_{s} \in \mathbb{R}^{d_z}$ corresponding to each point cloud $\bm{\mathcal{S}}$ is generated from the encoder.
For the cases where multiple latents represent different uniformly sampled variations of the same geometry, only the mean of the latent vectors is calculated and used. Hence, each geometry has its own unique latent vector.
For a geometry with $N_{\text{train}}$ boundary condition samples, this latent vector is first replicated
\begin{equation}
\mathcal{Z}_{s} = 
\begin{bmatrix}
\bm{z}_s & \dots &  \bm{z}_s
\end{bmatrix}^{T}
\in \mathbb{R}^{N_{\text{train}} \times d_z}.
\end{equation}
For each point cloud $\bm{\mathcal{S}}$, we load:
\begin{equation}
\mathcal{B}_{s} = 
\begin{bmatrix}
\bm{x}^{(1)}_{s} & \dots & \bm{x}^{(N_{\text{train}})}_{s}
\end{bmatrix}^{T}
\in \mathbb{R}^{N_{\text{train}} \times d}, \quad \mathcal{C}_{s}  = 
\begin{bmatrix}
\bm{y}^{(1)}_{s} & \dots & \bm{y}^{(N_{\text{train}})}_{s}
\end{bmatrix}^{T}
\in \mathbb{R}^{N_{\text{train}} \times m}
\end{equation}
where $\mathcal{B}_{s}$ is the boundary condition, $\mathcal{C}_{s}$ the target conditions (noisy systolic and diastolic pressure, and mean flow going through the left iliac artery and right iliac artery), and $\mathcal{Z}_{s}$ is the latent embedding for a single anatomy.
All anatomies are then concatenated, resulting in the training dataset $\mathcal{B}=\cup_{s=1}^{N}\mathcal{B}_{s}, \mathcal{C}=\cup_{s=1}^{N}\mathcal{C}_{s}$ and $\mathcal{Z}=\cup_{s=1}^{N}\mathcal{Z}_{s}$ with total number equal to $N\cdot N_{\text{train}}$ and dimensions equal to $d,m,d_z$, respectively.
Next, standardization is applied to the three datasets
\begin{equation}
\widetilde{\mathcal{B}} = \frac{\mathcal{B} - \mu_B}{\sigma_B}, \quad \widetilde{\mathcal{C}} = \frac{\mathcal{C} - \mu_C}{\sigma_C}, \quad \widetilde{\mathcal{Z}} = \frac{\mathcal{Z} - \mu_Z}{\sigma_Z}.
\end{equation}
where $\mu_B,\mu_C,\mu_Z$ and $\sigma_B, \sigma_C, \sigma_Z$ are the mean and standard deviations for $\mathcal{B}, \mathcal{C}$ and $\mathcal{Z}$ in the training data, respectively. 
Note that these statistics are reused for the validation and test datasets.
We train the CFM velocities $\bm{u}_t^{\bm{\theta}}(\bm{X}_{t}|\bm{Y},\bm{Z})$ from $\widetilde{\mathcal{B}}, \widetilde{\mathcal{Z}}, \widetilde{\mathcal{C}}$, parameterized by a multilayer perceptron with $d=6$, $d_z=6$, and $m=4$ for the model conditioned on both the geometry and the clinical targets. More details on the hyperparameters are in~\cref{app:optuna_hyperparams}.
We perform $k=6$ fold cross-validation over the set of geometries. For each fold, the framework partitions the dataset into training and validation, computes normalization constants based on the training examples, trains the model up to 5000 epochs, and applies early stopping with patience 500 and tolerance $10^{-4}$.
Let $\mathcal{L}_k^\ast$ denote the best validation loss for fold $k$. The cross-validation performance is summarized as
\begin{equation}
\overline{\mathcal{L}} = \frac{1}{K} \sum_{k=1}^{K} \mathcal{L}_k^\ast, \qquad \sigma_{\mathcal{L}} = \sqrt{\frac{1}{K-1}\sum_{k=1}^{K}\,\left(\mathcal{L}_k^\ast - \overline{\mathcal{L}}\right)^{2}}.
\end{equation}
This results in a latent-conditioned generative model capable of predicting conditional flow fields across geometries using the latent vectors produced by the encoder.

% ==============================================
\section{Numerical examples}\label{sec_examples}
% ==============================================

% ============================================
\subsection{Validation} \label{sec:validation}
% ============================================

As a first test case, we validated the CFM framework on the aorto-iliac bifurcation example from~\cite{choi2025_aobif}, which will also be used as the \emph{template} anatomy in later sections.

We generated a data set of 1000 low-fidelity points to train and test the CFM method. We added a noise of $\bm{\Sigma} = \text{diag}((0.05\,P_{\text{dia}})^2, (0.05\,P_{\text{sys}})^2)$ to the corresponding diastolic and systolic pressures. We split the data into 90\%-10\% training-testing. The specific hyperparameters used for this example can be found in \cref{app:optuna_hyperparams}.

\begin{figure}[!htb]
    \centering
    \includegraphics[width=\linewidth]{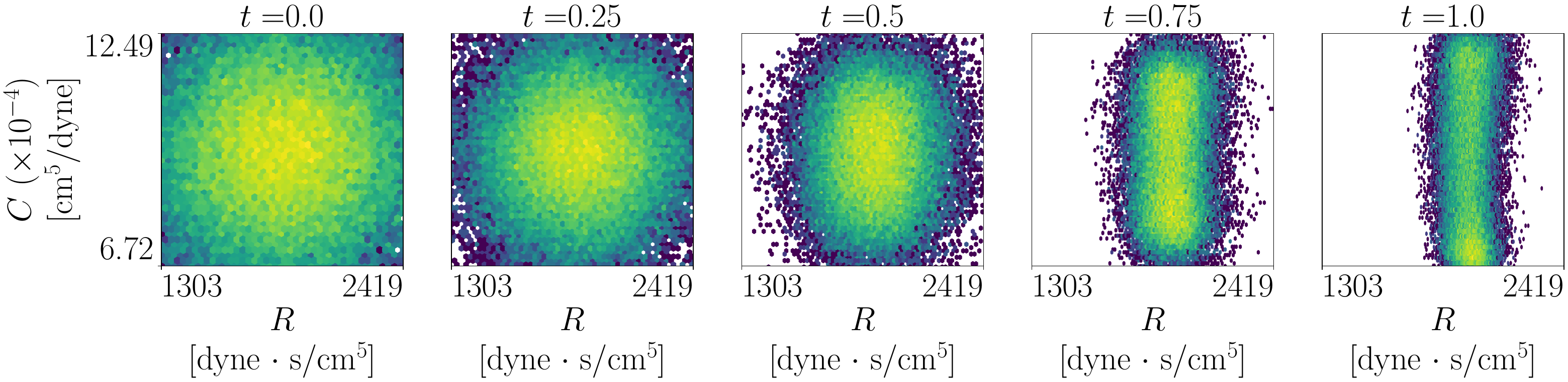}
    \caption{Posterior distribution from CFM over time $t \in [0,1]$.}
    \label{fig:CFM_aobif_time}
\end{figure}

\begin{figure}[!htb]
    \centering
    \includegraphics[width=0.5\linewidth]{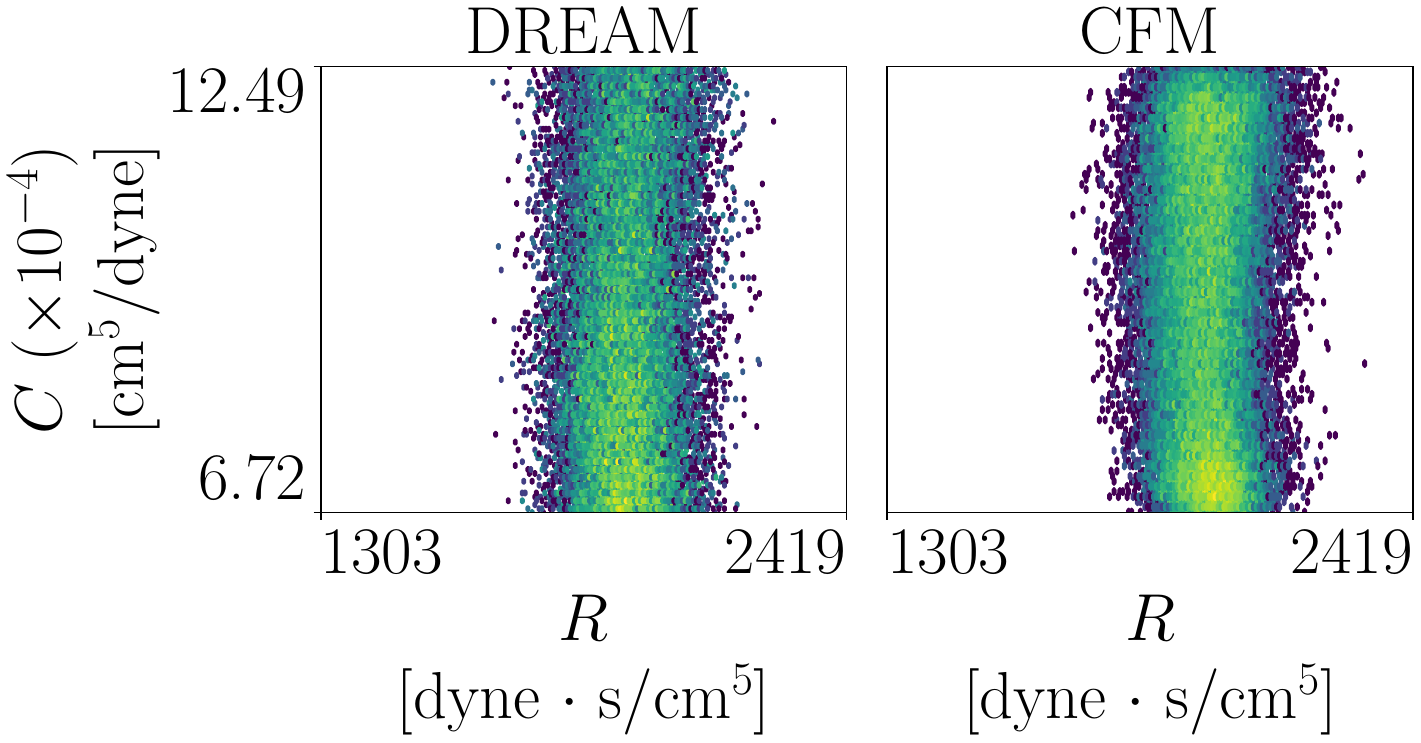}
    \caption{Samples from the DREAM and CFM posterior distributions for the aorto-iliac bifurcation example.}
    \label{fig:HF_validation}
\end{figure}

We chose one point from the offline testing set to visualize \cref{fig:CFM_aobif_time,fig:HF_validation}, corresponding to model outputs $P_{\text{dia}}=80$ mmHg and $P_{\text{sys}}=130$ mmHg, with noise $\mathcal{N}(0,\bm{\Sigma})$ as defined earlier added to them. Using this condition, we generated 50000 samples from the posterior distribution of the RC boundary conditions, as shown in \cref{fig:CFM_aobif_time}. 
When the auxiliary parameter $t$ varies from $0$ to $1$, the samples converge to the desired distribution, which coincides with that obtained in the aforementioned paper, showing the lack of identifiability (and therefore the lack of importance) of the capacitance boundary condition.
We compare the posterior distribution obtained from CFM with the posterior distribution from running differential evolution adaptive Metropolis (DREAM~\cite{vrugt_2009}) using the zero-dimensional solver in \cref{fig:HF_validation}.

\begin{figure}[!htb]
    \centering
    \includegraphics[width=0.75\linewidth]{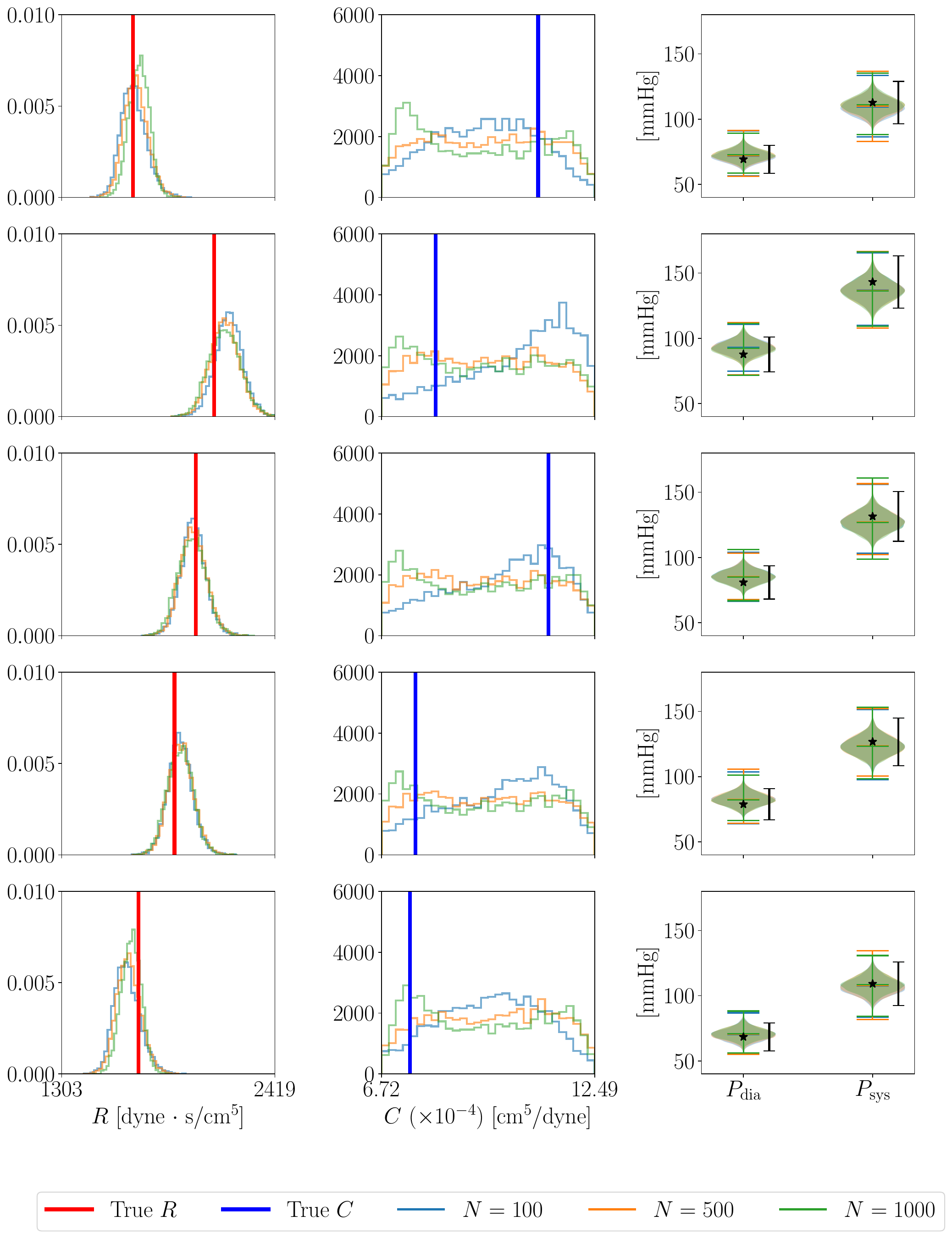}
    \caption{Single RC BCs: Results of posteriors for five different offline validation points (each row corresponds to a unique observation) subsampled from 10 offline testing data points. In the rightmost column, the true observations and the associated $\pm3$ standard deviations are displayed as a black star and a black bar, respectively.}
    \label{fig:lf_amortization_N100}
\end{figure}

% interpolation to other observations
After validation, we used the trained CFM to predict the posterior distributions for new clinical targets (observations), without retraining. \cref{fig:lf_amortization_N100} shows estimated densities from 5000 boundary condition realizations $\widehat{\bm{x}}$ generated by CFM, in addition to the true underlying boundary conditions.
In the same figure (right-most column) we also compare the posterior predictive distributions obtained from the zero-dimensional solver outputs corresponding to the estimated CFM boundary condition samples, with the distribution of the noisy observations. 
Summary statistics are finally reported in \cref{tab:summary_stats_2dims}. 
As noted for the previous test case, the marginal posterior for the capacitance boundary condition exhibits a large variance resulting from a lack of structural identifiability.
Furthermore, since CFM-based posteriors are generated for different observations without re-training, the figure also characterizes accuracy under amortization over unseen target pressures and flow split.
All predictive posterior distributions are closely centered around their corresponding truth, and mean reconstruction errors decrease with the increase in the training dataset size.

\begin{table}[!ht]
\centering
\caption{Single RC BC: Summary statistics of reconstruction error for $N=100, 500, 1000$. Absolute and signed errors are reported in mmHg. Relative errors are shown in parentheses (\%).}
\resizebox{0.8\linewidth}{!}{
\begin{tabular}{l c c}
\toprule
{\bf $N=100$} & \bf{$P_{\text{dia}}$ (mmHg, \%)} & \bf{$P_{\text{sys}}$ (mmHg, \%)} \\
\midrule
{\bf mean (abs/sign)} 
& 4.93 / 2.25 \; (6.34 / 3.26) 
& 6.59 / -1.58 \; (5.34 / -1.08) \\

{\bf std (abs/sign)} 
& 3.64 / 4.96 
& 4.84 / 7.11 \\
\midrule

{\bf $N=500$} & \bf{$P_{\text{dia}}$ (mmHg, \%)} & \bf{$P_{\text{sys}}$ (mmHg, \%)} \\
\midrule
{\bf mean (abs/sign)} 
& 4.95 / 1.00 \; (6.48 / 2.01) 
& 6.52 / -0.67 \; (5.50 / -0.43) \\

{\bf std (abs/sign)} 
& 3.50 / 4.86 
& 4.74 / 7.02 \\
\midrule

{\bf $N=1000$} & \bf{$P_{\text{dia}}$ (mmHg, \%)} & \bf{$P_{\text{sys}}$ (mmHg, \%)} \\
\midrule
{\bf mean (abs/sign)} 
& 4.22 / 0.90 \; (5.65 / 1.39) 
& 6.32 / 0.76 \; (5.59 / 0.93) \\

{\bf std (abs/sign)} 
& 3.09 / 4.52 
& 4.55 / 6.62 \\
\bottomrule
\end{tabular}}
\label{tab:summary_stats_2dims}
\end{table}

% ==========================================================
\subsection{Aorto-iliac bifurcation model with increasing numbers of boundary conditions}\label{sec:bifurcation_BCs}
% ==========================================================

In this section, we investigated the posterior distributions for three different cases: (1) total resistance and capacitance (two parameters in total), (2) total resistance and capacitance for each branch while keeping the $R_p/R_d$ ratio fixed (four parameters in total), and (3) proximal resistance, distal resistance, and capacitance per branch (six parameters in total).

Nominal values for the boundary conditions were obtained using Nelder-Mead optimization and a squared loss, considering clinical targets for the systolic and diastolic systemic pressure, equal to 120 mmHg and 80 mmHg, respectively. For Case (1), parameter point-estimates were obtained as $R_{\text{tot}} = 1861$ $\text{dynes } \cdot \text{s/cm}^5$, $C_{\text{tot}} = 9.61 \times 10^{-4}$ cm$^5$/dyne, and proximal-to-distal ratio of $8.84 \times 10^{-2}$. Results for Case (1) were tabulated and visualized in the previous section. The nominal values for Case (2) are listed in \cref{tab:aobif_baseline_4dims}, and Case (3) \cref{tab:aobif_baseline_6dims}.
Note that the flow split was assumed to be 42\% to the left iliac artery, based on the ratio of the cross-sectional area of the two iliac branches. This was a tunable parameter for Case (2) and Case (3).
We perform amortized inference using $N=100,500,1000$ low-fidelity data points, corresponding to the same aorto-iliac bifurcation model from~\cite{choi2025_aobif}, as before. 
We added heteroscedastic noise with distribution $\mathcal{N}(\bm{0},\bm{\Sigma})$, where $\bm{\Sigma} = \text{diag}( (0.05\, P_{\text{dia}})^2, (0.05\,P_{\text{sys}})^2)$, and $P_{\text{dia}}$, $P_{\text{sys}}$ the diastolic and systolic pressure, respectively. 
We used 10\% of the data points for offline testing, and split the remaining data into 75\%-25\% online training and validation. 
Hyperparameters corresponding to this section are listed in \cref{app:optuna_hyperparams}.
We list the reconstruction errors and posterior distributions for Case (2) in \cref{sec:4_BC_params}.

\begin{table}[!ht]
    \centering
    \caption{Nominal values for the two RCR BCs with fixed ratios (four dimensions, two per branch outlet) for aorto-iliac model.}
    \resizebox{0.5\linewidth}{!}{
    \begin{tabular}{l c c}
    \toprule
    {\bf Vessel} & {\bf $R_{\text{tot}}$ [$\text{dynes}\cdot\text{s}/\text{cm}^5$]} & {\bf $C$ [cm$^5$/dyne]}\\
    \midrule
     {\bf Right iliac} & 3228 & 5.949$\times$10$^{-4}$\\
     {\bf Left iliac} & 4399 & 3.678$\times$10$^{-4}$ \\
         \bottomrule
    \end{tabular}}
    \label{tab:aobif_baseline_4dims}
\end{table}

\begin{figure}[!htb]
    \centering
    \includegraphics[width=\linewidth]{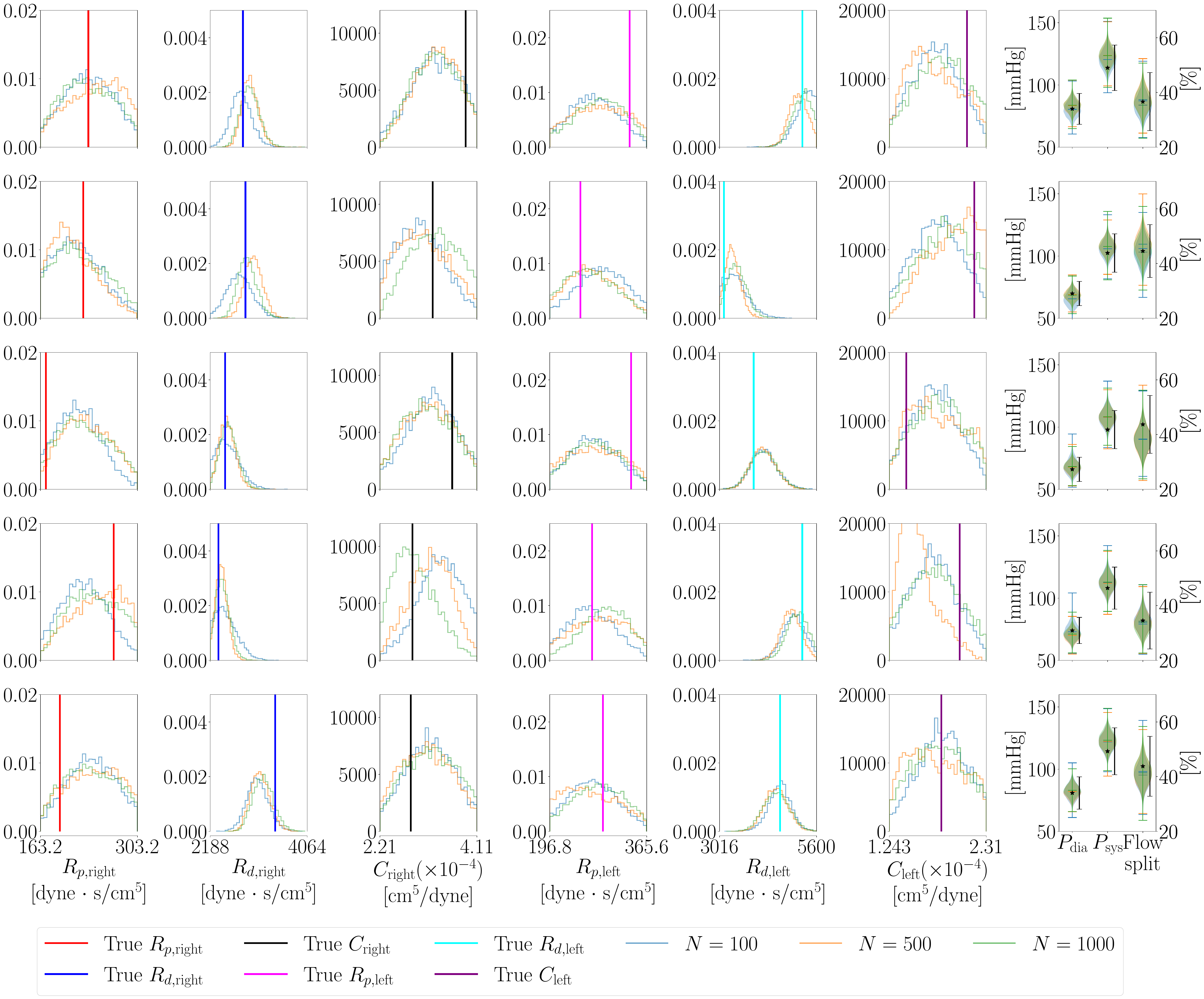}
    \caption{Two RCR BCs (6 dimensions): Results of marginal posteriors for five offline validation points subsampled from 100 offline testing data points. In the rightmost column, true observations and corresponding $\pm3$ standard deviations are illustrated with a black star and a black bar, respectively.}
    \label{fig:lf_amortization_6dim}
\end{figure}

We logged the absolutely reconstruction errors in \cref{tab:summary_stats_6dim} and obtained the marginal posterior distributions in \cref{fig:lf_amortization_6dim}. Qualitatively, we see that all the predictive posterior distributions are closely centered around the true observation. Quantitatively, we see that the mean reconstruction errors decrease as more samples are added to the expected noise level, i.e. 4.39 mmHg for diastolic pressure, 6.84 mmHg for systolic pressure, and 3.91\% flow split.

\begin{table}[!ht]
    \centering
    \caption{Nominal values for aorto-iliac model with two RCR BCs (6 dimensions)}
    \resizebox{0.7\linewidth}{!}{
    \begin{tabular}{l c c c}
    \toprule
    {\bf Vessel} & {\bf $R_p$ [$\text{dynes}\cdot\text{s}/\text{cm}^5$]} & {\bf $C$ [cm$^5$/dyne]} & {\bf $R_d$ [$\text{dynes}\cdot\text{s}/\text{cm}^5$]}\\
    \midrule
     {\bf Right iliac} & 233.2 & 3.161$\times$10$^{-4}$ & 3126 \\
     {\bf Left iliac} & 281.2 & 1.775$\times$10$^{-4}$ & 4308 \\
         \bottomrule
    \end{tabular}}
    \label{tab:aobif_baseline_6dims}
\end{table}

\begin{table}[!ht]
\centering
\caption{Two RCR BCs: Summary statistics for $N=100, 500, 1000$ data. Absolute reconstruction errors are shown. Relative errors are reported in parentheses (\%).}
\resizebox{\linewidth}{!}{
\begin{tabular}{l c c c }
\toprule
{\bf $N=100$} & \bf{$P_{\text{dia}}$ (mmHg, \%)} & \bf{$P_{\text{sys}}$ (mmHg, \%)} & \bf{Flow split (\% flow, \% rel)} \\
\midrule
{\bf mean (abs/sign)} 
& 5.31 / -1.59 \; (6.92 / -1.99) 
& 7.52 / 0.80 \; (6.47 / 1.08) 
& 3.98 / -0.24 \; (9.38 / 0.19) \\

{\bf std (abs/sign)} 
& 3.89 / 6.00 
& 5.30 / 7.34 
& 2.92 / 4.36 \\
\midrule

{\bf $N=500$} 
& \bf{$P_{\text{dia}}$ (mmHg, \%)} & \bf{$P_{\text{sys}}$ (mmHg, \%)} & \bf{Flow split (\% flow, \% rel)} \\
\midrule
{\bf mean (abs/sign)} 
& 4.25 / 1.07 \; (5.69 / 1.68) 
& 6.88 / -0.97 \; (5.76 / -0.42) 
& 3.67 / 0.02 \; (8.79 / 0.53) \\

{\bf std (abs/sign)} 
& 3.16 / 4.81 
& 4.75 / 6.72 
& 2.73 / 4.19 \\
\midrule

{\bf $N=1000$} 
& \bf{$P_{\text{dia}}$ (mmHg, \%)} & \bf{$P_{\text{sys}}$ (mmHg, \%)} & \bf{Flow split (\% flow, \% rel)} \\
\midrule
{\bf mean (abs/sign)} 
& 4.39 / 0.74 \; (5.81 / 1.30) 
& 6.84 / 1.14 \; (5.98 / 1.36) 
& 3.91 / 0.13 \; (9.57 / 1.22) \\

{\bf std (abs/sign)} 
& 3.24 / 4.85 
& 4.79 / 6.75 
& 2.87 / 4.23 \\
\bottomrule
\end{tabular}}
\label{tab:summary_stats_6dim}
\end{table}

% ==============================================================================
\subsection{Aorto-iliac model with two RCR BCs (6 dimensions) plus parametric inflow (19 dimensions), heartbeat, and fundamental frequency}\label{sec:RCR_bc_inflow}
\begin{figure}[!htb]
    \centering
    \includegraphics[width=\linewidth]{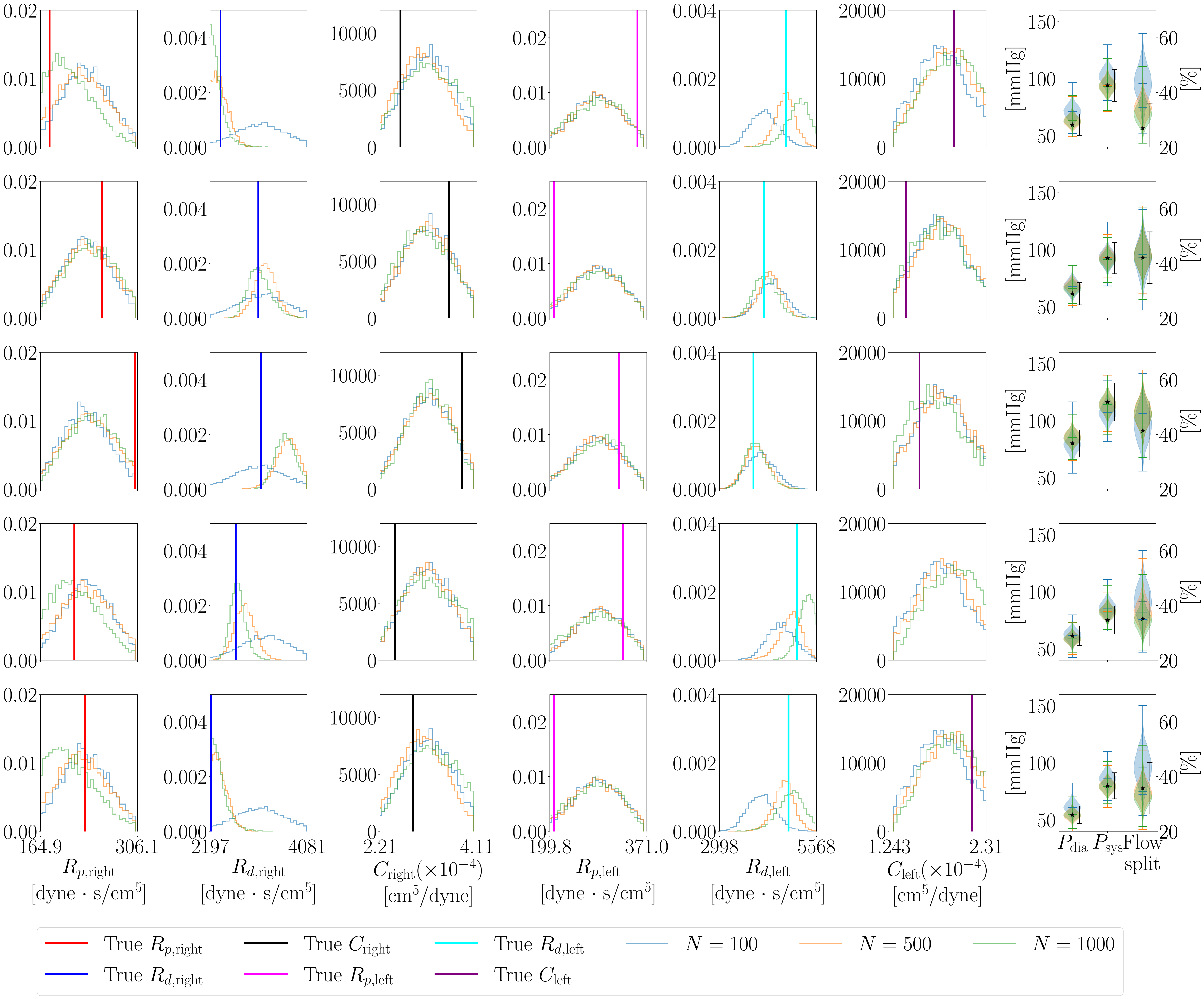}
    \caption{Marginal distributions of RCR BCs (6 dimensions, 3 per outlet), conditioned on parametric inflow, target pressures and flow split. 
    Results of marginal posteriors for five offline validation samples (one per row). 
    In the rightmost column, the true pressures and flow split are illustrated with a black star,  with $\pm3$ standard deviations in black bars.}
    \label{fig:marginals_6dims_conditioned_on_inflow}
\end{figure}

% ==============================================================================
%
Recall that a periodic function $f(t)$ with fundamental period $T$ (i.e., heart cycle duration), can be expressed through a Fourier series of the form
\begin{equation}
   f(t) = a_0 + \sum_{n=1}^{\infty} \left[ a_n \cos\left(\frac{2\pi n t}{T} \right) + b_n \sin \left(\frac{2\pi n t}{T} \right) \right],
\end{equation}
where the Fourier coefficients are given by
\begin{equation}
    a_0 = \frac{1}{T} \int_0^T f(t)dt, \quad a_n = \frac{2}{T} \int_0^T f(t) \cos\left(\frac{2\pi n t}{T} \right) dt, \quad b_n = \frac{2}{T} \int_0^T f(t) \sin\left(\frac{2\pi n t}{T} \right) dt.
\end{equation}
We retained the mean, $a_0$, and the first nine pairs of Fourier coefficients to parametrize the aortic flow profile. 
We used the inflow profiles of 16 abdominal aorta models from the Vascular Model Repository (VMR)~\cite{vmr} to train a flow matching-based estimator from their Fourier features, so we could generate an arbitrary number of inflow profiles to compile training data (see \cref{app:inflows} for additional details). 
We then used this set of inflows as \emph{conditions} to the CFM framework, in addition to the target pressures and mean flows.
FalconBC then estimated the proximal resistance, distal resistance, and capacitance.
\cref{fig:marginals_6dims_conditioned_on_inflow} shows how the predictive posterior distribution progressively approximates the true observations and associated aleatoric uncertainty when the number of samples increases.
Additionally, \cref{tab:summary_stats_aobif_6dims_inflow} shows a corresponding reduction of approximation errors. 
Since the dimensionality of the conditioning set increases sensibly with the addition of the inflow features, changes in summary statistics due to an increasing training dataset size are more apparent than in the previous sections.

\begin{table}[!ht]
\centering
\caption{Amortized inference of two RCR BCs (one per outlet) conditioned on parametric inflow. Absolute errors statistics for training dataset size $N=100, 500, 1000$. Relative errors are reported in parentheses (\%).}
\resizebox{\linewidth}{!}{
\begin{tabular}{l c c c }
\toprule
{\bf $N=100$} & \bf{$P_{\text{dia}}$ (mmHg, \%)} & \bf{$P_{\text{sys}}$ (mmHg, \%)} & \bf{Flow split (\% flow, \% rel)} \\
\midrule
{\bf mean (abs/sign)} 
& 9.08 / 1.23 \; (13.23 / 3.82) 
& 8.48 / 0.93 \; (8.65 / 2.11) 
& 7.26 / 1.38 \; (19.14 / 6.79) \\

{\bf std (abs/sign)} 
& 5.11 / 6.33 
& 5.66 / 7.22 
& 4.37 / 5.63 \\
\midrule

{\bf $N=500$} 
& \bf{$P_{\text{dia}}$ (mmHg, \%)} & \bf{$P_{\text{sys}}$ (mmHg, \%)} & \bf{Flow split (\% flow, \% rel)} \\
\midrule
{\bf mean (abs/sign)} 
& 4.47 / 0.89 \; (6.90 / 2.08) 
& 6.78 / -0.99 \; (6.63 / -0.42) 
& 4.35 / -0.14 \; (10.51 / 0.98) \\

{\bf std (abs/sign)} 
& 3.23 / 4.72 
& 4.46 / 6.10 
& 3.01 / 4.24 \\
\midrule

{\bf $N=1000$} 
& \bf{$P_{\text{dia}}$ (mmHg, \%)} & \bf{$P_{\text{sys}}$ (mmHg, \%)} & \bf{Flow split (\% flow, \% rel)} \\
\midrule
{\bf mean (abs/sign)} 
& 4.30 / -0.33 \; (6.17 / -0.07) 
& 6.15 / -0.56 \; (6.09 / -0.17) 
& 3.79 / -0.72 \; (8.81 / -1.14) \\

{\bf std (abs/sign)} 
& 3.11 / 4.53 
& 4.27 / 5.97 
& 2.77 / 4.13 \\
\bottomrule
\end{tabular}}
\label{tab:summary_stats_aobif_6dims_inflow}
\end{table}

\subsubsection{Joint estimation of Fourier coefficients and boundary conditions}

\begin{figure}[!htb]
    \centering
    \includegraphics[width=0.5\linewidth]{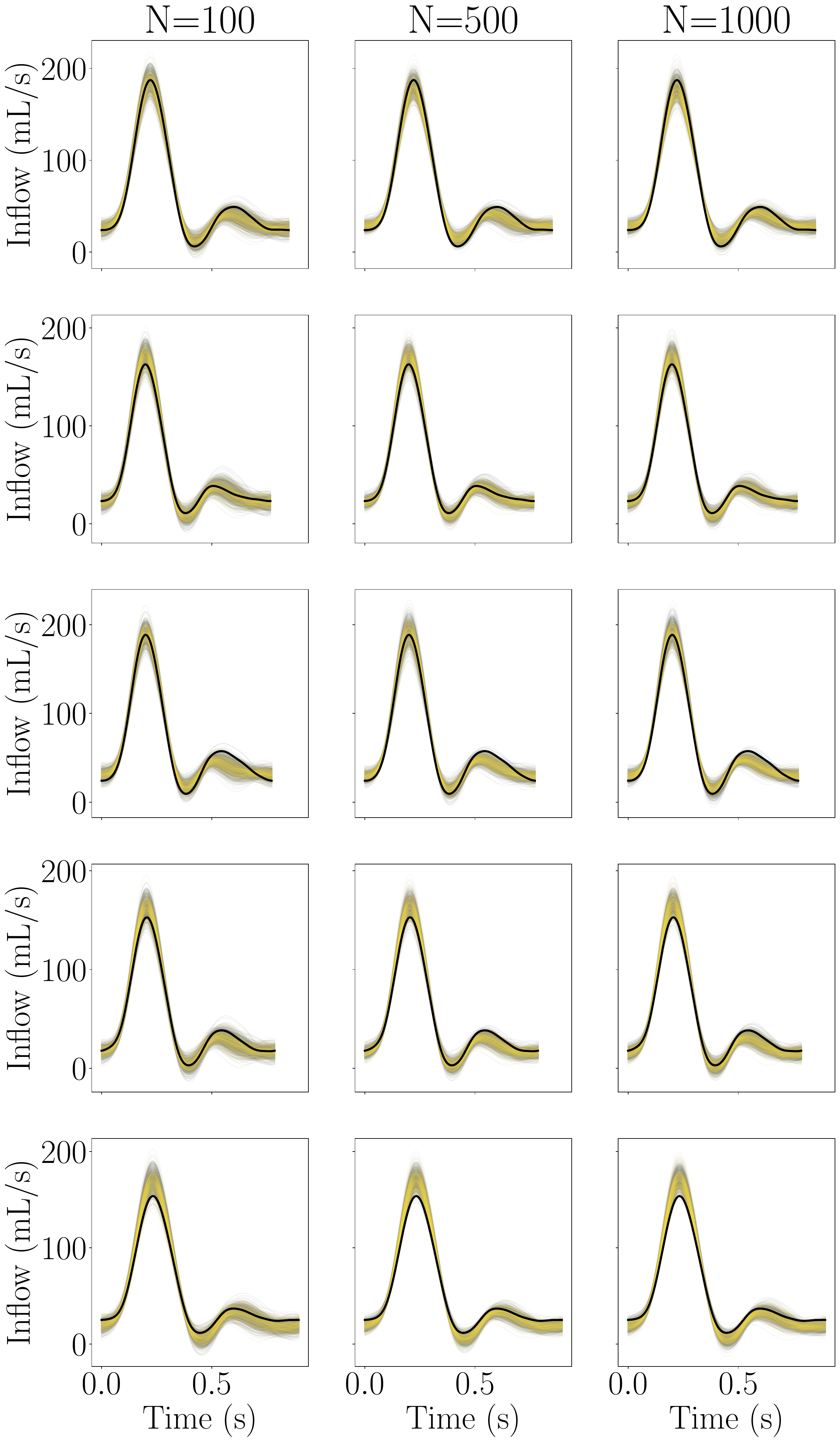}
    \caption{Two RCR BCs and Fourier coefficients: Results of 1000 estimated Fourier curves subsampled from the posterior distribution. True curve is in black.}
    \label{fig:predicted_fourier_coeffs}
\end{figure}

We further add details on joint estimation of the Fourier coefficients and the boundary conditions in \cref{app:geometry_inference}. We show that the Fourier coefficients corresponding to the cardiac output and fundamental frequency can be inferred from the CFM framework in \cref{fig:predicted_fourier_coeffs}.

% =========================================================================
\subsection{Conditioning on model anatomy}\label{sec:geometry_conditioning}
% =========================================================================

We finally introduce two different approaches to create interpretable latent representations of diseased anatomies for varying stenosis severity. 
We condition CFM on these latent representations plus noisy pressure and flow split targets, and generate conditional distributions (posterior) of boundary condition parameters.
We also curate a testing dataset consisting of the following anatomies unseen at training: Location A with 56.7\% stenosis, Location B with 66.7\% stenosis, Location C with 60.3\% stenosis, Location D with 69.6\% stenosis, Location E with 68.9\% stenosis, and Location F with 60.9\% stenosis from \cref{fig:stenosis_suite}. 

% =====================================================
\subsubsection{User-specified one-hot anatomy encoding}
% =====================================================

\begin{figure}[!htb]
    \centering
    \includegraphics[width=1.0\linewidth]{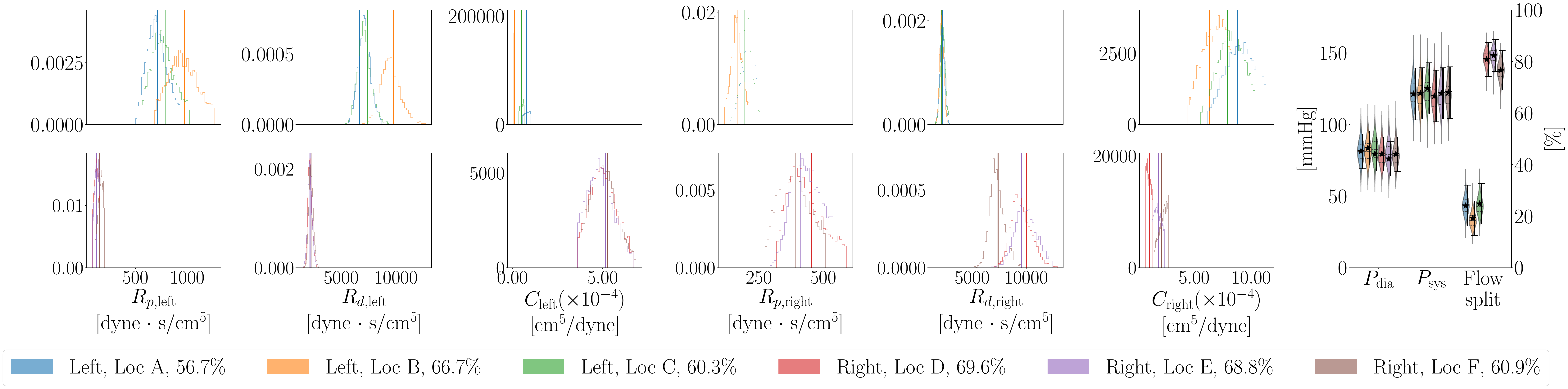}
    \caption{Marginal posteriors and corresponding true boundary conditions at testing. In the rightmost column, true observations and corresponding $\pm3$ standard deviations are illustrated with black stars and black bars, respectively.}
    \label{fig:user_defined_latent_histograms}
\end{figure}

\begin{figure}[!htb]
    \centering
    \includegraphics[width=\linewidth]{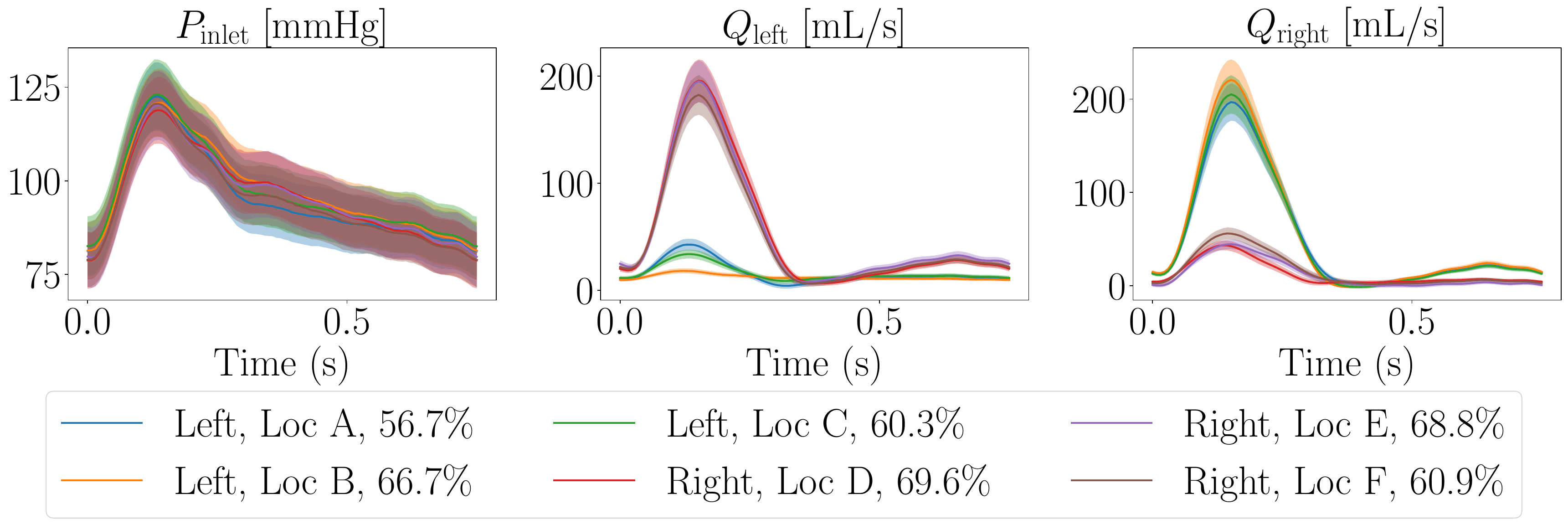}
    \caption{Posterior predictive time traces of inlet pressure and outlet flows. Shaded regions indicate $\pm$ one standard deviation.}
    \label{fig:user_defined_latent_fwd}
\end{figure}

Consider a user-defined six-dimensional latent space, where each component corresponds to one of the six locations in \cref{fig:stenosis_suite}, with magnitude equal to degree of stenosis at that location. 
This effectively establishes a one-to-one mapping from the stenosis location and severity to the entire diseased anatomy. 
We then used the user-defined latents and $N=1000$ input-output pairs to train the CFM framework with $k$-fold cross-validation. 
The tuned hyperparameters are tabulated in \cref{tab:appendix_optuna_hyperparams}. 
We reserved $N=6$ unseen geometries for testing, with results shown in \cref{fig:user_defined_latent_histograms}. 
\cref{fig:user_defined_latent_fwd} shows the posterior predictive distributions of inlet pressure and outlet flows, obtained by forward propagating the estimated CFM conditions through the zero-dimensional solver.

%========================================================
\subsubsection{Encoding-decoding anatomical point clouds}
%========================================================

\begin{figure}[!htb]
    \centering
    \includegraphics[width=1.0\linewidth]{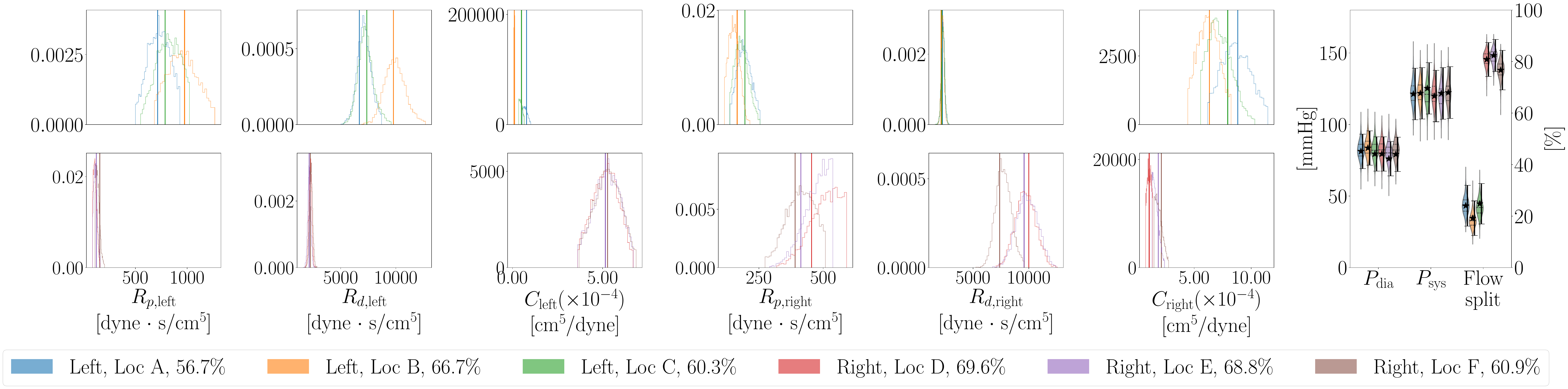}
    \caption{Encoding-decoding anatomical point clouds. Marginal posterior distributions at testing. In the rightmost column, true underlying systolic, diastolic pressures and flow split and corresponding $\pm3$ standard deviations are illustrated with a black star and a black bar, respectively.}
    \label{fig:encoder_histograms}
\end{figure}

\begin{figure}[!htb]
    \centering
    \includegraphics[width=\linewidth]{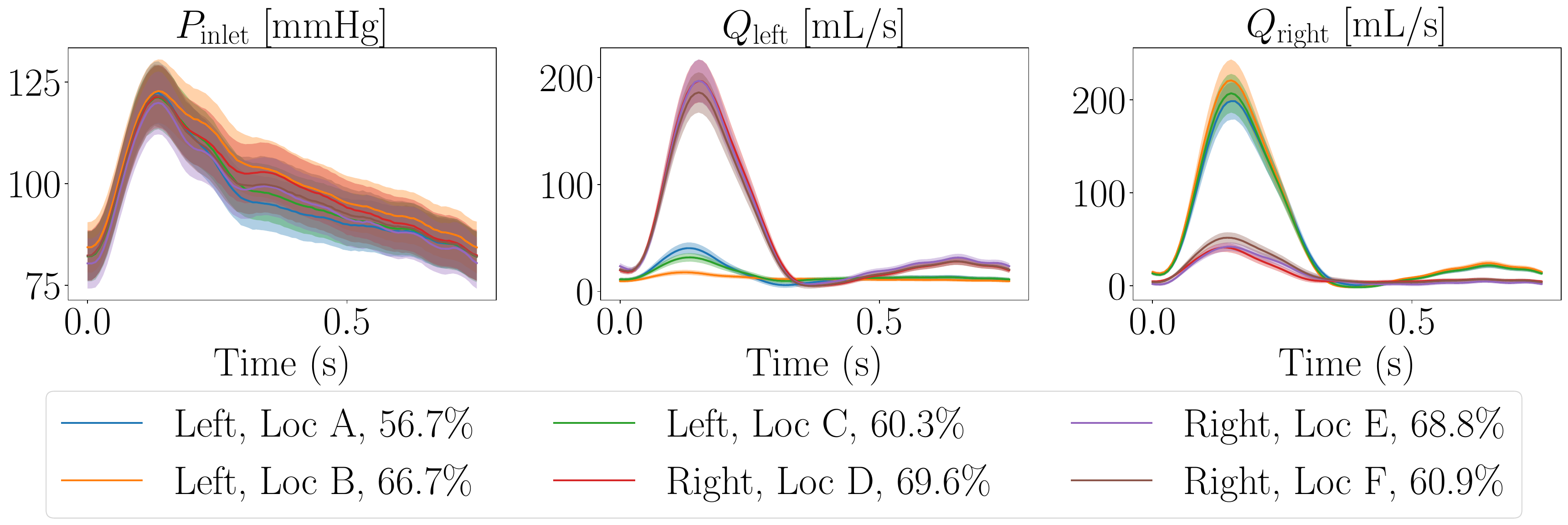}
    \caption{Encoding-decoding anatomical point clouds. Posterior predictive time traces of inlet pressure and outlet flows. Shaded regions indicate $\pm$ one standard deviation.}
    \label{fig:encoder_fwd}
\end{figure}

\begin{figure}[!htb]
    \centering
    \includegraphics[width=\linewidth]{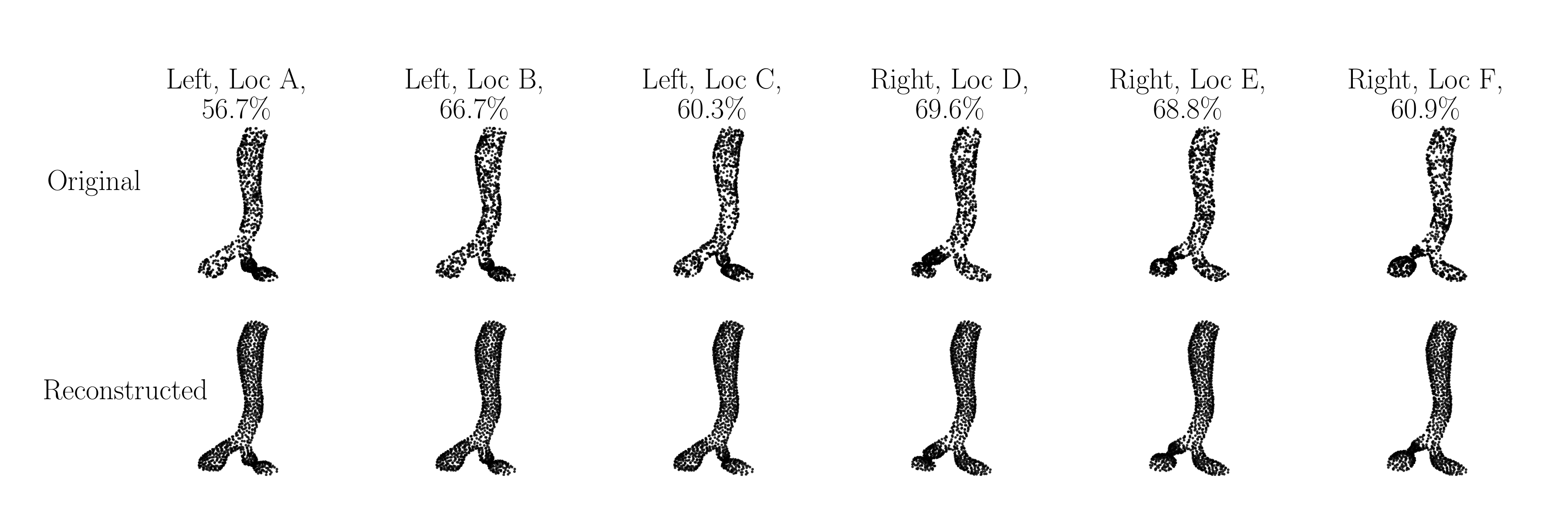}
    \caption{Reconstructed geometries using the complete encoder-decoder architecture in \cref{fig:encoderdecoder}.}
    \label{fig:encoder_geometries}
\end{figure}

% geometry conditioning
This section considers an alternative mechanism to produce an embedding from a diseased anatomy. Given a reference point cloud and a diseased point cloud, the encoder learns stenosis severity and corresponding location. 
The decoder then uses the latent vector constructed from this information to reconstruct the geometry. 
After the encoder-framework was trained for 5000 epochs, the latent vectors generated by the encoder were used to train the CFM model for another 5000 epochs with 6-fold cross validation.
Tuned hyperparameters are reported in \cref{tab:appendix_optuna_hyperparams}.
The resulting marginal posteriors and corresponding true boundary conditions are shown in \cref{fig:encoder_histograms}, whereas the posterior predictive time traces for the inlet pressures and outlet flows are displayed in \cref{fig:encoder_fwd}.
Finally, we show the reconstructed geometries that were recovered from this new latent space in \cref{fig:encoder_geometries}.

\begin{remark}[Selection of seemingly equivalent RCR BCs]
{\it To generate the training dataset for CFM, our \emph{prior} consisted of uniform perturbations from a nominal set of optimized RCR boundary conditions. We initially tuned to the same clinical targets, which resulted in flow reversal at the outlet of the left iliac artery, in cases with significant stenosis of the right iliac artery.
To mitigate this issue, we determined the flow split from the optimization, increased parameter ranges, and included additional constraints to the differential evolution optimization in \cref{app:geometry_bc_optimization}.
We believe this to be an interesting case where observing the model response in different scenarios helped to mitigate clinical target reachability and parameter non-identifiability.}
\end{remark}

% ==================================================================
\subsubsection{Joint estimation of geometry and boundary conditions}
% ==================================================================

\begin{figure}[!htb]
    \centering
    \includegraphics[width=\linewidth]{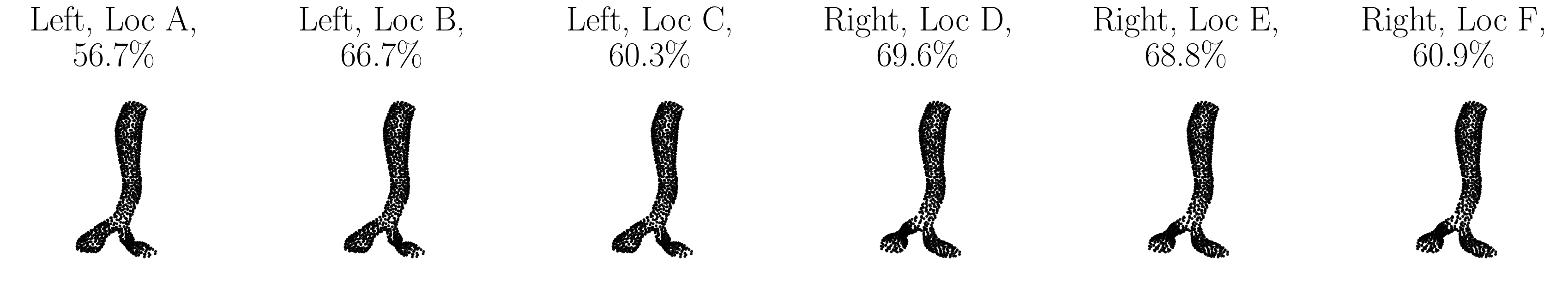}
    \caption{Joint estimation of anatomical point clouds and boundary conditions from CFM. The point clouds are obtained by decoding the marginal posterior samples for the anatomical latent variables, representing stenosis location and severity.}
    \label{fig:reconstructed_geoms_from_estimation}
\end{figure}

We add further details on joint geometry and boundary condition estimation in \cref{app:geometry_inference}. We display the posterior distributions for the boundary conditions obtained from the CFM model in the appendix. Here, we plot the reconstructed geometries in \cref{fig:reconstructed_geoms_from_estimation}, which were obtained by passing the inferred latents to the trained decoder. We observe that the \emph{unsupervised} CFM model learns a six-dimensional latent representation that naturally separates into two approximately three-dimensional subspaces: one corresponding to the left iliac artery and one corresponding to the right iliac artery. This separation arises because the zero-dimensional simulations depend strongly on which iliac artery contains the stenosis. A stenosis in the left iliac artery alters the effective resistance of the blood vessel on the left, while a stenosis in the right iliac artery alters the resistance of the blood vessel on the right. These changes induce structurally distinct simulation outputs, since the lumped-parameter model computes blood vessel-averaged pressures and flow rates. As a result, geometries with left-sided stenosis and right-sided stenosis occupy different regions in output space. The CFM model, which is trained to match these simulation outputs, therefore learns to embed them in well-separated regions of latent space.

In contrast, the specific axial location of the stenosis along a given branch has a much weaker effect on the zero-dimensional simulation results, as expected. 
The learned latent representation primarily captures which artery is affected and the overall severity of the stenosis, while variations in location are weakly expressed.

% ===============================================================
\subsection{Towards broader clinical application: coronary artery disease}\label{sec:coronary}
% ===============================================================

We have thoroughly investigated various strategies for boundary condition tuning in peripheral artery disease using a model of the aorto-iliac bifurcation.
We devote this subsection to discuss some broader clinical implications of our work.

\begin{figure}[htb!]
    \label{fig:coronary}
    \centering
    \begin{subfigure}[b]{\textwidth}
        \centering
        \includegraphics[width=0.5\linewidth]{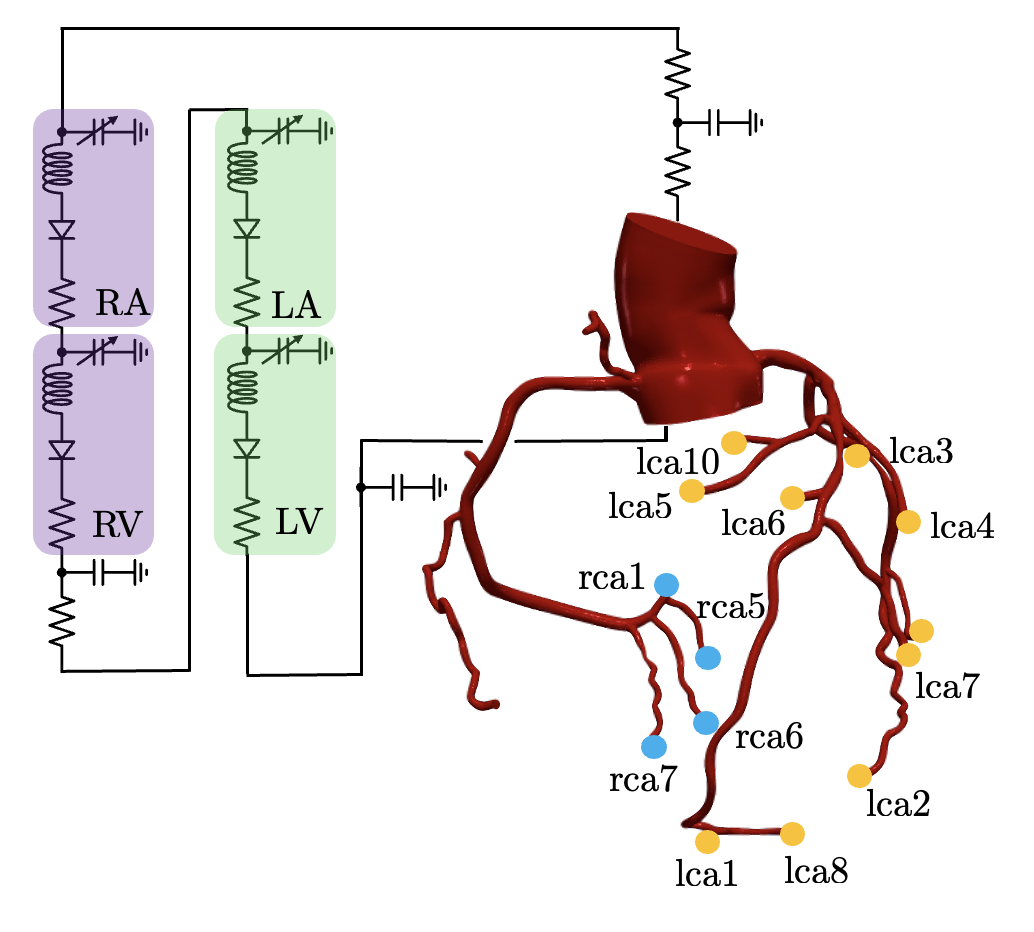}
        \caption{Visualization of coronary model with partially labeled left (lca) and right (rca) coronary artery outlet BCs. 
        The four-chamber boundary condition model (closed-loop) consists of the left ventricle (LV), left atrium (LA), right ventricle (RV), right atrium (RA), and a systemic and pulmonary circulation, represented through a RCR and RC circuit, respectively.}
        \label{fig:coronary_3d}
    \end{subfigure}
    
    \vspace{5mm}
    
    \begin{subfigure}[b]{\textwidth}
        \centering
        \includegraphics[width=0.65\linewidth]{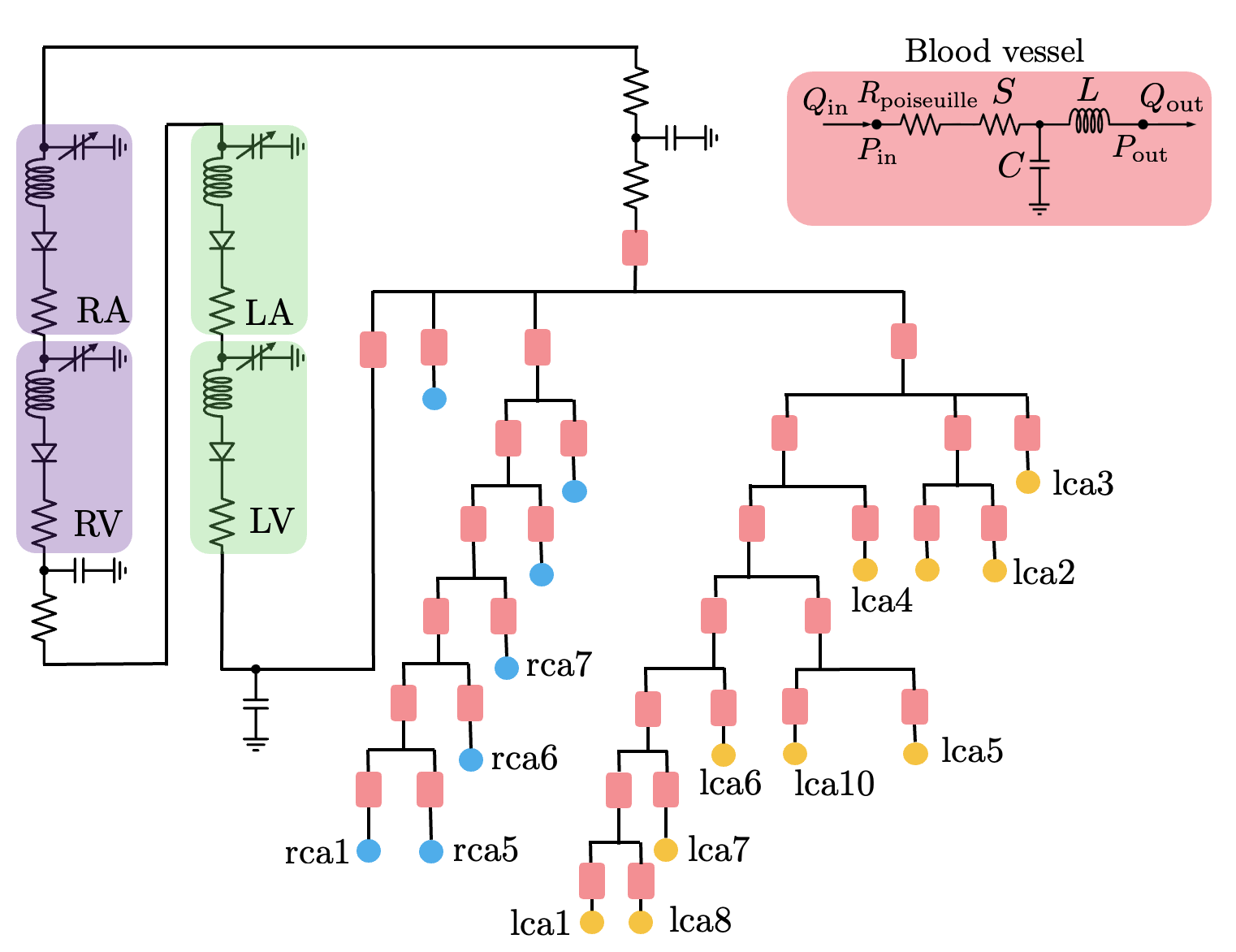}
        \caption{Schematic of the zero-dimensional coronary model.}
        \label{fig:coronary_0d}
    \end{subfigure}
    
    \vspace{5mm}
    
    \begin{subfigure}[b]{\textwidth}
        \centering
        \includegraphics[width=0.5\linewidth]{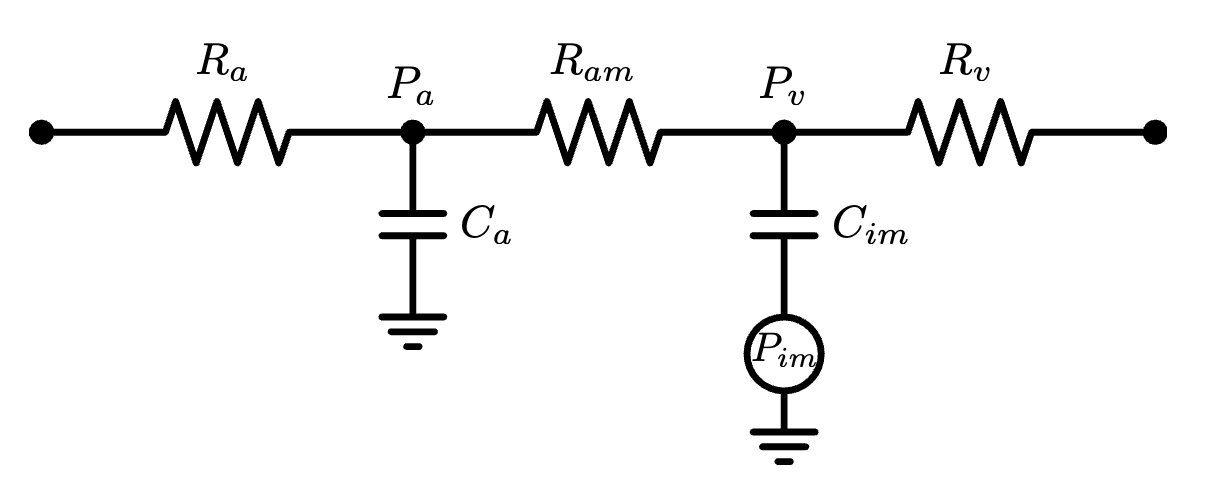}
        \caption{Circuit model for lca and rca outlet boundary conditions, applied to the yellow and blue locations in (b).}
        \label{fig:coronary_outlet_bcs}
    \end{subfigure}
    \caption{Coronary model and closed-loop boundary conditions.}
\end{figure}

As our final example, we consider a 14-dimensional boundary condition tuning exercise with a zero-dimensional patient-specific model relevant to the diagnosis and treatment of coronary artery disease (CAD) from~\cite{MZK24}.
Blood is assumed Newtonian with density $\rho=1.06$ g/cm$^3$ and dynamic viscosity $\mu=$ 0.04 dynes/cm$^2$.
We considered closed-loop boundary conditions, and used a lumped parameter (0D) circulation model which includes the four heart chambers, accounts for pulmonary resistance, and considers a Windkessel boundary condition at the aortic outlet, as shown in \cref{fig:coronary_3d} and \cref{fig:coronary_0d}.
Additionally, we applied coronary outflow boundary conditions and included a distal resistance $R_{a}+R_{am}+R_{v}$, with prescribed intramyocardial pressure $P_{im}$ \cite{kim2010patient,sankaran2012patient,tran16_automated}, as indicated with blue and yellow circles in \cref{fig:coronary_outlet_bcs}.

We followed the problem set-up in \cite{MZK24} and used FalconBC in the final step of parameter estimation, to infer the resistance $\bm{r}$ at the coronary outlets.
In total, we consider $N_c=14$ random variables, $\bm{x}=\{r_1, r_2, ..., r_{N_c}\}$, one for each coronary artery perfusing the left ventricle, as shown in \cref{fig:coronary_3d}.
We infer outlet resistances from target flow rates $\bm{y}_{CT} \in \mathbb{R}^{N_c}$ measured through a new technique known as myocardial perfusion imaging~\cite{nieman2020dynamic}. 
%
%We use a non-diagonal covariance matrix estimated from 2500 realizations of myocardial blood flow (MBF) data (see Sec. 2.2 in \cite{MZK24} for details). 
%
A uniformly distributed prior within $[0.5\,r_{i},2.0\,r_{i}]$ is used for each coronary resistance, where the nominal value $r_{i}$, $i=1,\dots,N_{c}$ is predetermined based on relative target flows for each artery.
After all the coronary resistances are determined, we multiply the total coronary mean flow by a scaling factor of $R_{\text{tot,truth}}/R_{\text{tot,predicted}}$ to preserve total coronary mean flow $R_{\text{tot,truth}}$ determined from a previous optimization step, where $R_{\text{tot}}=(\sum_{i=1}^n 1/r_i)^{-1}$.

\begin{figure}[!htb]
    \centering
    \includegraphics[width=\linewidth]{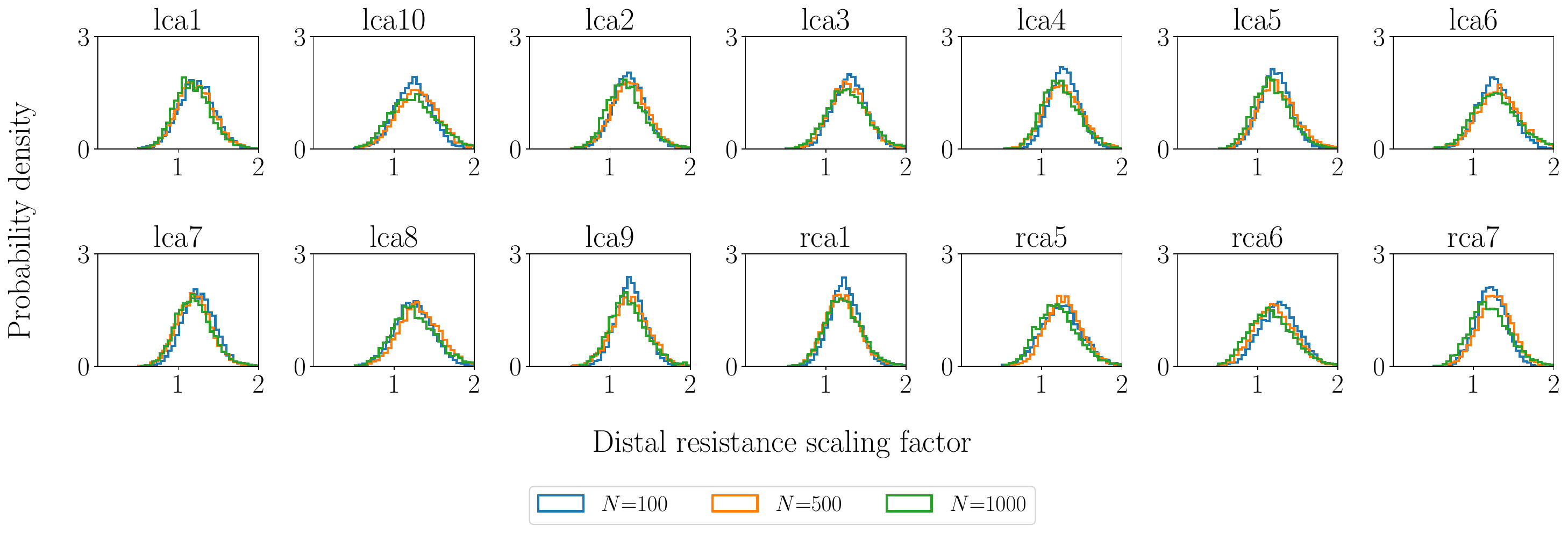}
    \caption{Marginal posterior distributions for the resistance scaling factors at the coronary outlets.}
    \label{fig:coronary_posteriors}
\end{figure}

\begin{figure}[!htb]
    \centering
    \includegraphics[width=\linewidth]{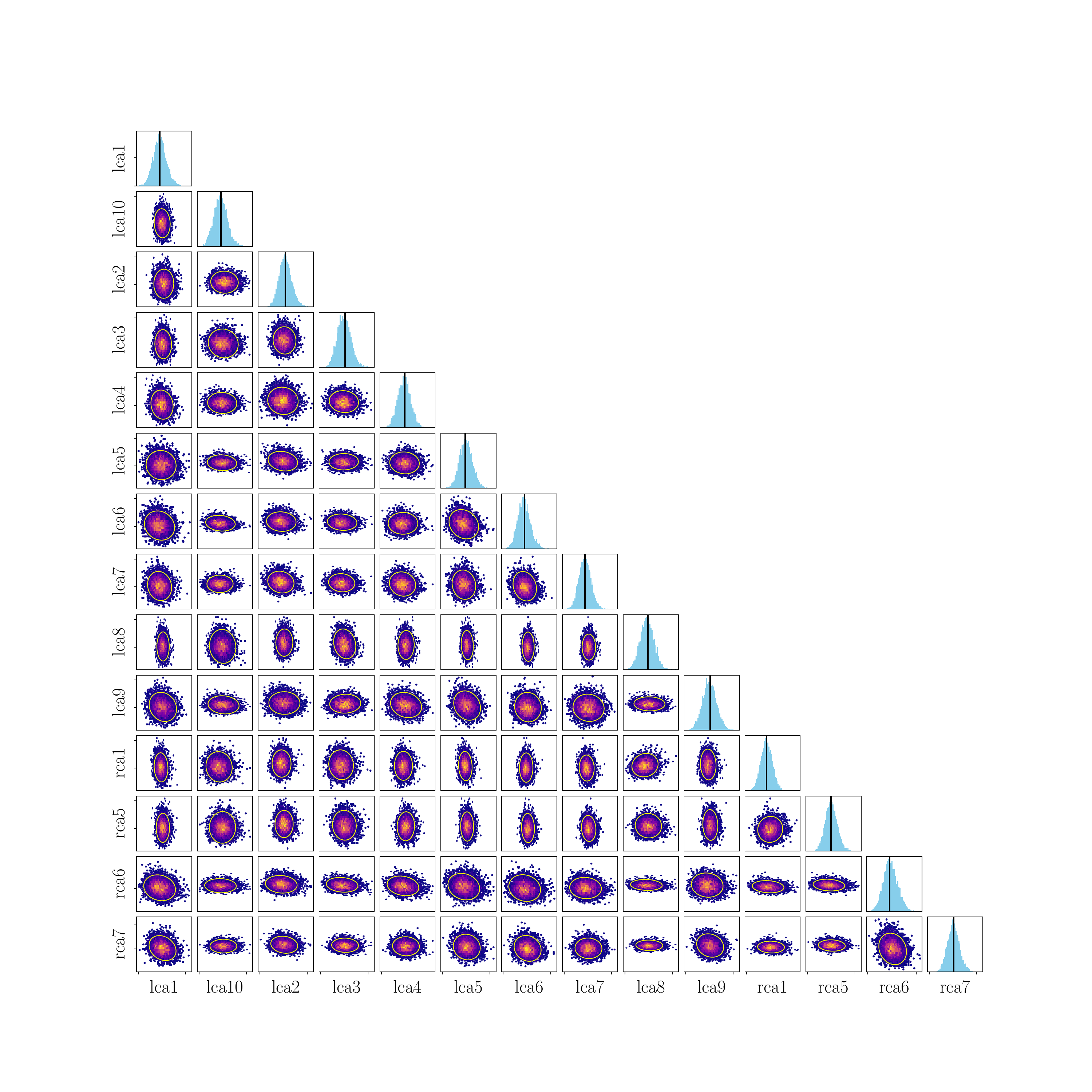}
    \caption{Distributions of simulated (hexbin) versus measured (yellow ellipse) univariate and bivariate marginal posterior predictive samples for branch-specific coronary flows computed by FalconBC.}
    \label{fig:coronary_results_dist}
\end{figure}

FalconBC was trained using $N=100, 500, 1000$ realizations. The hyperparameters leading to minimal testing losses are tabulated in \cref{tab:appendix_hyperparameters_MLP} for each training dataset size. Estimated posteriors in \cref{fig:coronary_posteriors} show concentration around 1.0, while the corresponding posterior predictive distributions in \cref{fig:coronary_results_dist} confirm a remarkable agreement with the measured covariance (yellow ellipses, showing $3 \sigma$ level sets), demonstrating the accuracy resulting from FalconBC.

In~\cite{MZK24}, 24 parallel Markov chains with a total of 10,000 generations each were run, taking 16.9 hours on 24 CPUs. In contrast, after offline training FalconBC, for which hyperparameter optimization takes 9 minutes for $N=100$ and 77 minutes for $N=1000$, it takes approximately 0.13 seconds on one CPU to produce the $N=5000$ posterior boundary condition samples (even after rejecting samples that are out of bounds).
This suggests that, once trained, FalconBC can be used to quickly retune a cardiovascular model to better match clinical targets.

% ========================================================
\section{Discussion and Conclusions}\label{sec:conclusion}
% ========================================================

In this study, we demonstrated the flexibility and accuracy of FalconBC for amortized inference of boundary conditions in cardiovascular modeling. 
Previously, much of the boundary condition tuning process in the cardiovascular community has been done by recursive (and often manual) evaluation of expensive models or Bayesian sampling approaches. However, each of these methods is costly, laborious and requires re-tuning given new observations.
In this work, we introduced a fast amortized inference framework that uses both simulation outputs and geometries to \textit{generate} new samples, i.e., we introduced the efficacy of a \textit{generative model} for boundary condition tuning. The \emph{generative learning trilemma}~\cite{xiao2021tackling} states that no generative model can simultaneously optimize all three of the following properties: (i) sample quality, (ii) diversity/coverage, and (iii) computational efficiency. In order to transition from traditional MCMC to generative modeling, we need a framework that addresses all of these points, so we aim to highlight the power of CFM for potential use for improving boundary condition tuning in cases where real-time inference is needed or in cases where different anatomies need to be incorporated. Although this work has demonstrated the methodology on one specific patient with varying levels and locations of stenosis, the elements in our overall proposed framework can be modified to incorporate anatomies of patients with different topologies, which we leave as future work.

In this work, we highlighted that different deformed anatomies and inflow curves could be used as \emph{conditions} to guide parameter estimation, or could be \emph{jointly estimated} with the boundary conditions to improve reachability of clinical targets.
% using inflow features
Specifically, we extended classical efforts in boundary condition tuning under fixed inflow conditions, with the ability \emph{to either condition on, or to jointly estimate inflow features}, e.g., Fourier coefficient representation of periodic inflow waveforms.
While traditional methods used averaged Fourier coefficients (e.g., in \cite{les10_aaa}) or statistical shape modeling methods (e.g., in \cite{saitta2023data}) to generate inflow curves, we showed that flow matching can also generate meaningful families of inflow waveforms. 
Flow matching-based generation can be used for scenarios in which the cardiac output (or relevant mean inflow value) for the patient is known, but no (e.g., MRI) data is available to characterize the entire waveform, or for cases in which the inflow waveform (e.g., inferior vena cava flow for Fontan patients) can be measured by MRI but is only a portion of the cardiac output. With sufficient data diversity, FalconBC can be used to \emph{estimate} a posterior distribution which jointly quantify boundary conditions and inflow waveform Fourier coefficients.
This goes well beyond existing practices of inflow scaling, providing an opportunity, for example, to quantify the correlation between inflow waveform features and population variability or demographic information.

% encoding anatomy
We showed how the same principle can be extended to features encoding an entire anatomical model, focusing on the extraction of a six-dimensional latent space embedding from a point cloud characterization of the lumen surface from an aorto-iliac bifurcation model.
This opens up to two new possibilities in boundary condition tuning. On the one hand, this allows for continuous data feedback under a changing geometry in a digital twin scenario (e.g., during transcatheter interventions), where, e.g., instead of re-tuning boundary conditions in real-time, one could adjust the parameters of a blood vessel. On the other hand, this allows for joint estimation of geometrical features and boundary conditions under uncertainty to improve the current patient-specific model at hand, given limited or noisy data.
The latter is useful, as sometimes the pressure measured with a catheter is not \emph{reachable} with boundary condition tuning due to imperfections in the segmented anatomy. This approach holds the potential to instead \emph{estimate} necessary changes in local anatomical features (e.g. diameter) that would lead to the desired pressures or flow splits, going well beyond current possibilities.

% use of a zero-dimensional model and appropriateness for aorto-iliac
Furthermore, the current study focuses on using zero-dimensional simulations to generate training data for FalconBC. While this is appropriate for the selected model, due to the close agreement shown in \cref{fig:3d_vs_0d_results}, it might lead to excessive approximation in cases, e.g., pulmonary models, with numerous branches, and more pronounced minor losses due to bifurcation, see, e.g., \cite{lee24_ppas}. This could be mitigated by combining FalconBC with recent multi-fidelity architectures that learn the correlations between high- and low-fidelity models, such as that introduced in \cite{meng2020composite,villatoro2025emulator}.

% rigid wall assumptions
We also assumed rigid wall in our study, but an extension to deformable walls would only affect the cost of generating the training dataset, with no other changes required for the FalconBC pipeline.
% effect of exercise: bimodal inflow...
Other interesting aspects include the effect of exercise at the presence of a stenosis, which could result in complicated flow phenomena. These effects may cause larger discrepancies in pressure drop predictions in 3D simulations that may not be accurately captured by 0D models.

Finally, while this work primarily focused on an aorto-iliac bifurcation model and probed the effect of various degrees of stenosis percentages and locations on the posterior distributions, we hope to incorporate larger patient cohorts with different topologies (e.g., different number of outlets) in future work. 

% ===========================
\subsection*{Acknowledgments}
% ===========================

CHC acknowledges support from the Yansouni Family Stanford Graduate Fellowship and AHA Predoctoral Fellowship 26PRE1550972.
ALM acknowledges support from NSF grant \#2105345. 
ALM and DES acknowledge support from NIH grant \#1R01HL167516 {\it Uncertainty aware virtual treatment planning for peripheral pulmonary artery stenosis}. 
High performance computing resources were provided by the San Diego Supercomputing Cluster and Sherlock High Performance Computing Cluster.

\bibliographystyle{mrp}
\bibliography{biblio}

\newpage
\appendix

\section{Hyperparameter tuning} \label{app:optuna_hyperparams}

The hyperparameters of the neural networks used in the numerical examples are tuned using 100 Optuna~\cite{Optuna} iterations on a single CPU. 
The dataset was split into 25\% testing and 75\% training samples. 
The hyperparameters included in the optimization procedure, along with their admissible ranges, are listed in \cref{tab:appendix_optuna_hyperparams}. 
In particular, we consider number of hidden layers $\ell$, number of neurons per hidden layer $n$, learning rate $h$, and mini-batch size $s$. All neural networks have $\mathrm{SiLU}$ activation function and are trained for 15,000 epochs using the Adam optimizer \cite{adam_opt_2017} and a \texttt{ExponentialLR()} scheduler with 0.9999. The hyperparameters selected via Optuna for the MLP model, which are used in the numerical experiments, are reported in \cref{tab:appendix_hyperparameters_MLP}. We track the best validation loss and save the associated model weights for later evaluation. Further note that, due to the small ($N_{\text{geom}}=48$) data set size of the geometry, the CFM ``One-hot'' and ``Encoder-decoder'' in \cref{tab:appendix_hyperparameters_MLP} were trained with $k$-fold cross validation, and the following bounds were used for weight decay tuning: $\{10^{-6},\dots,10^{-4}\}$. 
The corresponding tuned hyperparameters for weight decay were: 9.81$\times$10$^{-5}$ for ``One-hot,'' 7.22$\times$10$^{-5}$ for ``Encoder-decoder,'' 2.77$\times$10$^{-6}$ for ``One-hot with geometry estimation,'' and 1.182$\times$10$^{-5}$ for ``Encoder-decoder with geometry estimation''.

\begin{table}[!ht]
    \centering
    \caption{Hyperparameters and corresponding admissible ranges tuned using Optuna.}
    \begin{tabular}{c c}
        \toprule
        {\bf Hyperparameter} & {\bf Range}\\
        \midrule
        Hidden layers $\ell$ & \{1, \dots,  4\} \\
        Neurons per hidden layer $n$ & \{1, \dots, 128\} \\
        Learning rate $h$ & [1e-4, 1e-2] \\
        Mini-batch size $s$ & \{8, \dots, 64\} for $N=100$ \\
        & \{16, \dots, 128\} for $N=500, 1000$ \\
        \bottomrule
    \end{tabular}
    \label{tab:appendix_optuna_hyperparams}
\end{table}

\begin{table}[!ht]
    \centering
    \caption{FalconBC hyperparameters selected via Optuna. The parameters are: number of hidden layers $\ell$, number of neurons per hidden layer $n$, learning rate $h$ and mini-batch size $s$.}
    \begin{tabular}{c|cccccc}
    \toprule
         Example & $\ell$ & $n$ & $h$ & $s$ \\
    \midrule
        Validation & 2 & 102 & 0.000431 & 32 \\ %& 1 & 110 & 0.0017284 & 24 \\
        Single RC $(N=100)$ & 2 & 7 & 0.0004859 & 16\\
        Single RC $(N=500)$ & 1 & 20 & 0.0001535 & 16\\
        Single RC $(N=1000)$ & 2 & 102 & 0.000431 & 32 \\
         Two RCRs with fixed $R_p/R_d$ $(N=100)$ & 1 & 106 & 0.0002250 & 16 \\
        Two RCRs with fixed $R_p/R_d$ $(N=500)$ & 1 & 30 & 0.0012486 & 16 \\
         Two RCRs with fixed $R_p/R_d$ $(N=1000)$ & 4 & 27 & 0.0001544 & 32 \\
        Two RCRs $(N=100)$ & 3 & 8 & 0.0001531 & 16 \\
        Two RCRs $(N=500)$ & 1 & 84 & 0.0002433 & 16 \\
        Two RCRs $(N=1000)$ & 1 & 80 & 0.0001494 & 32 \\
        Two RCRs with inflow condition $(N=100)$ & 3 & 1 & 0.0004929 & 32 \\
        Two RCRs with inflow condition $(N=500)$ & 3 & 3 & 0.0005434 & 16 \\
        Two RCRs with inflow condition $(N=1000)$ & 3 & 11 & 0.0046388 & 32 \\
        Two RCRs with inflow estimation $(N=100)$ & 1 & 49 & 0.0001014 & 64 \\
        Two RCRs with inflow estimation $(N=500)$ & 1 & 48 & 0.0007291 & 16 \\
        Two RCRs with inflow estimation $(N=1000)$ & 1 & 48 & 0.0018371 & 32 \\
        14 coronary outlets ($N=100$) & 2 & 128 & 0.00013214 & 8\\
        14 coronary outlets ($N=500$) & 2 & 30 & 0.0077286 & 32 \\
        14 coronary outlets ($N=1000$) & 2 & 31 & 0.0024792 & 32 \\
        One-hot & 1 & 100 & 0.0045622 & 16 \\
        Encoder-decoder & 4 & 20 & 0.0005042 & 32 \\
        One-hot with geometry estimation & 3 & 89 & 0.0004522 & 64 \\
        Encoder-decoder with geometry estimation & 3 & 109 & 0.0001095 & 64 \\
    \bottomrule    
    \end{tabular}
    \label{tab:appendix_hyperparameters_MLP}
\end{table}

% ==========================================================
\section{Aorto-iliac bifurcation model with increasing numbers of boundary conditions}
% ==========================================================

% ============================================================================
\subsection{Aorto-iliac model with two RCR BCs and fixed ratios (4 dimensions)}\label{sec:4_BC_params}
\begin{figure}[!htb]
    \centering
    \includegraphics[width=\linewidth]{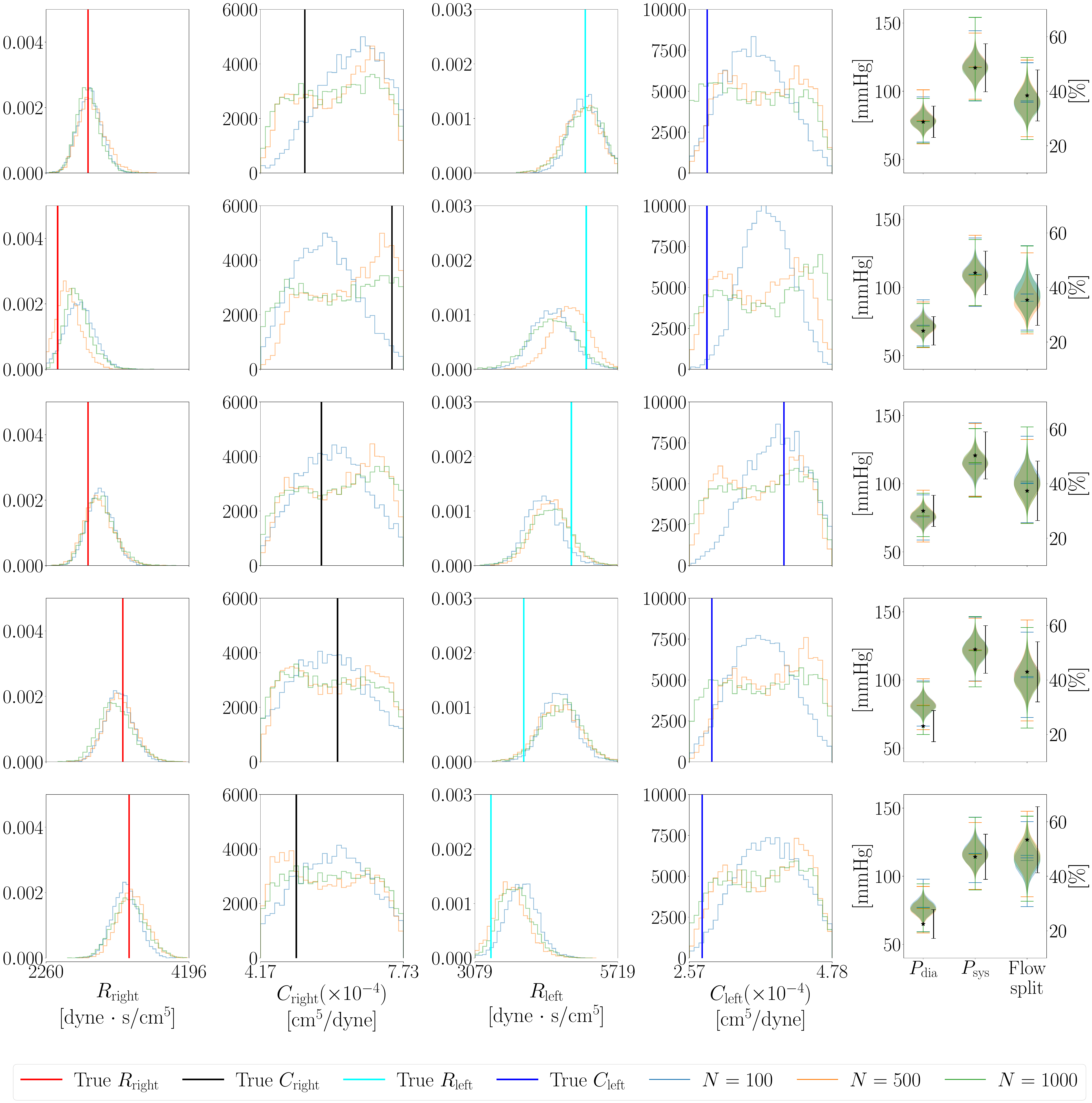}
    \caption{Two RC BCs: Results of marginal posteriors for five offline validation points subsampled from 50 offline testing data points. In the rightmost column, true observations and corresponding $\pm3$ standard deviations are illustrated with a black star and a black bar, respectively.}
    \label{fig:lf_amortization_4dim}
\end{figure}

% ============================================================================
%
We determined optimal boundary conditions (see \cref{tab:aobif_baseline_4dims}) from a Nelder-Mead optimizer, specifying clinical target pressures equal to 120/80 mmHg, and left/right flow split of 42\%/58\%. We imposed RCR boundary conditions to both branches, and fixed the proximal-to-distal ratios to be $8.84 \times 10^{-2}$ for each iliac artery.
We created varying data set sizes of $N=100, 500, 1000$ using only a zero-dimensional cardiovascular models with found boundary condition parameters. 
We showcased the results in \cref{fig:lf_amortization_4dim}, where we see that the mean of the distributions overlaps with the target observation values. 
We logged the actual range, mean, and standard deviation of the reconstruction errors in \cref{tab:summary_stats_4dims}. Qualitatively, we see that all the predictive posterior distributions are closely centered around the true observation. Quantitatively, we see that the mean (signed) reconstruction errors decrease as more samples are added to 0.79 mmHg for diastolic pressure, 0.96 mmHg for systolic pressure and 0.19\% flow split.

\begin{table}[!ht]
    \centering
    \caption{Nominal values for the two RCR BCs with fixed ratios (four dimensions, two per branch outlet) for aorto-iliac model.}
    \begin{tabular}{l c c}
    \toprule
    {\bf Vessel} & {\bf $R_{\text{tot}}$ [$\text{dynes}\cdot\text{s}/\text{cm}^5$]} & {\bf $C$ [cm$^5$/dyne]}\\
    \midrule
     {\bf Right iliac} & 3228 & 5.949 $\times 10^{-4}$\\
     {\bf Left iliac} & 4399 & 3.678 $\times 10^{-4}$ \\
         \bottomrule
    \end{tabular}
    \label{tab:aobif_baseline_4dims}
\end{table}

\begin{table}[!ht]
\centering
\caption{Two RC BCs: Summary statistics for training dataset size $N=100, 500, 1000$. Absolute errors are reported in mmHg. Relative errors are shown in parentheses (\%).}
\resizebox{\linewidth}{!}{
\begin{tabular}{l c c c }
\toprule
{\bf $N=100$} & \bf{$P_{\text{dia}}$ (mmHg, \%)} & \bf{$P_{\text{sys}}$ (mmHg, \%)} & \bf{Flow split (\% flow, \% rel)} \\
\midrule
{\bf mean (abs/sign)} 
& 5.50 / 2.97 \; (7.53 / 4.49) 
& 6.64 / 0.41 \; (5.71 / 0.63) 
& 3.71 / -0.54 \; (8.31 / -0.77) \\

{\bf std (abs/sign)} 
& 3.53 / 4.76 
& 4.72 / 6.88 
& 2.77 / 4.21 \\
\midrule

{\bf $N=500$} 
& \bf{$P_{\text{dia}}$ (mmHg, \%)} & \bf{$P_{\text{sys}}$ (mmHg, \%)} & \bf{Flow split (\% flow, \% rel)} \\
\midrule
{\bf mean (abs/sign)} 
& 4.59 / 1.10 \; (5.95 / 1.70) 
& 7.01 / 0.92 \; (5.90 / 1.09) 
& 3.86 / 0.04 \; (9.22 / 0.91) \\

{\bf std (abs/sign)} 
& 3.35 / 4.89 
& 4.94 / 6.98 
& 2.87 / 4.26 \\
\midrule

{\bf $N=1000$} 
& \bf{$P_{\text{dia}}$ (mmHg, \%)} & \bf{$P_{\text{sys}}$ (mmHg, \%)} & \bf{Flow split (\% flow, \% rel)} \\
\midrule
{\bf mean (abs/sign)} 
& 4.24 / 0.79 \; (5.59 / 1.30) 
& 6.79 / 0.96 \; (5.88 / 1.13) 
& 3.85 / 0.19 \; (9.54 / 1.31) \\

{\bf std (abs/sign)} 
& 3.11 / 4.62 
& 4.72 / 6.74 
& 2.82 / 4.26 \\
\bottomrule
\end{tabular}}
\label{tab:summary_stats_4dims}
\end{table}

% ============================================================================
\subsection{Aorto-iliac model conditioned on clinical targets, CO, and fundamental frequency to estimate two RCR BCs (6 dimensions) and 10 pairs of Fourier coefficients}\label{sec:6_BC_and_fourier_estimation}
\begin{figure}[!htb]
    \centering
    \includegraphics[width=\linewidth]{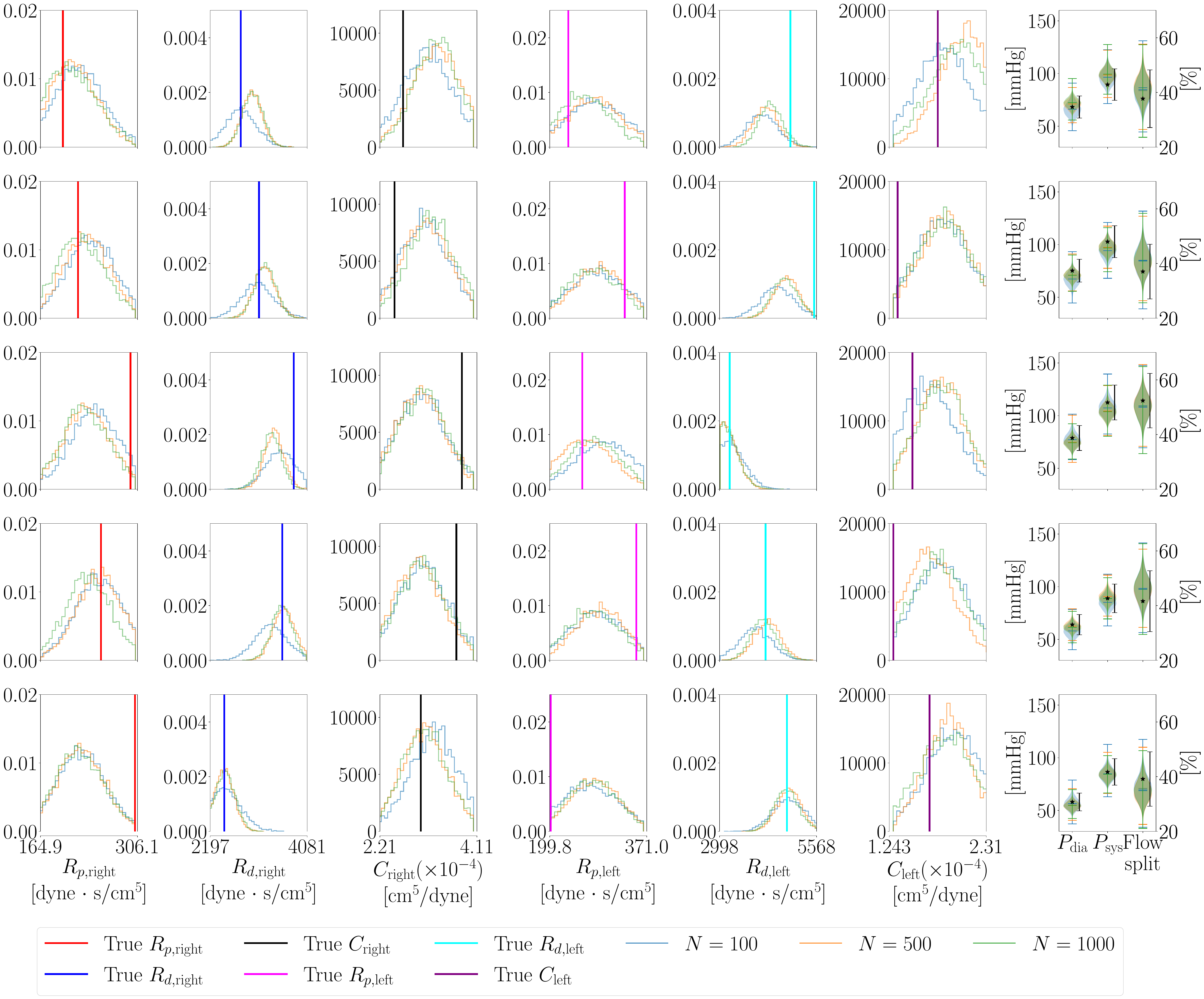}
    \caption{Joint estimation of two RCR BCs and inflow Fourier features: Results of marginal posteriors for five offline validation points subsampled from 50 offline testing data points. In the rightmost column, true observations and corresponding $\pm3$ standard deviations are illustrated with a black star and a black bar, respectively.}
    \label{fig:lf_amortization_6dims_with_fourier}
\end{figure}

% ============================================================================
%
We determined optimal boundary conditions from a Nelder-Mead optimizer, specifying clinical target pressures equal to 120/80 mmHg, left/right flow splits, and cardiac output and fundamental frequency. We imposed RCR boundary conditions to both branches.
We created varying data set sizes of $N=100, 500, 1000$ using a zero-dimensional cardiovascular model, leading to an input parameter space with twenty six dimensions.
We showcased the results in \cref{fig:lf_amortization_6dims_with_fourier}, where we see that the mean of the distributions overlaps with the target observation values. 
We logged the actual range, mean, and standard deviation of the reconstruction errors in \cref{tab:summary_stats_6dims_fourier_estimation}. All the predictive posterior distributions are closely centered around the true observation, with mean (signed) reconstruction errors decreasing, as more samples are added, to 0.42 mmHg for diastolic pressure, 0.09 mmHg for systolic pressure and 0.38\% flow split.

\begin{table}[!ht]
\centering
\caption{Joint estimation of two RCR BCs and inflow Fourier features: Summary statistics for training dataset size $N=100, 500, 1000$. Absolute errors are reported in mmHg. Relative errors are shown in parentheses (\%).}
\resizebox{\linewidth}{!}{
\begin{tabular}{l c c c }
\toprule
{\bf $N=100$} & \bf{$P_{\text{dia}}$ (mmHg, \%)} & \bf{$P_{\text{sys}}$ (mmHg, \%)} & \bf{Flow split (\% flow, \% rel)} \\
\midrule
{\bf mean (abs/sign)} 
& 6.65 / -1.16 \; (9.85 / -0.95) 
& 8.46 / -1.85 \; (8.17 / -0.84) 
& 3.85 / -0.16 \; (9.16 / 0.19) \\

{\bf std (abs/sign)} 
& 4.54 / 6.41 
& 5.43 / 7.39 
& 2.91 / 4.45 \\
\midrule

{\bf $N=500$} 
& \bf{$P_{\text{dia}}$ (mmHg, \%)} & \bf{$P_{\text{sys}}$ (mmHg, \%)} & \bf{Flow split (\% flow, \% rel)} \\
\midrule
{\bf mean (abs/sign)} 
& 4.62 / 0.32 \; (6.75 / 0.92) 
& 6.00 / 0.46 \; (6.06 / 0.83) 
& 3.81 / 0.18 \; (9.45 / 1.25) \\

{\bf std (abs/sign)} 
& 3.27 / 4.65 
& 4.25 / 6.08 
& 2.82 / 4.18 \\
\midrule

{\bf $N=1000$} 
& \bf{$P_{\text{dia}}$ (mmHg, \%)} & \bf{$P_{\text{sys}}$ (mmHg, \%)} & \bf{Flow split (\% flow, \% rel)} \\
\midrule
{\bf mean (abs/sign)} 
& 4.00 / 0.42 \; (6.07 / 1.10) 
& 5.46 / 0.09 \; (5.60 / 0.32) 
& 4.12 / 0.38 \; (10.34 / 2.17) \\

{\bf std (abs/sign)} 
& 2.95 / 4.44 
& 4.01 / 5.97 
& 2.95 / 4.21 \\
\bottomrule
\end{tabular}}
\label{tab:summary_stats_6dims_fourier_estimation}
\end{table}

%===============================================================
\section{Aorto-iliac model estimation on model geometry}\label{app:geometry_inference}
% % ===============================================================

\begin{figure}[!htb]
    \centering
    \includegraphics[width=\linewidth]{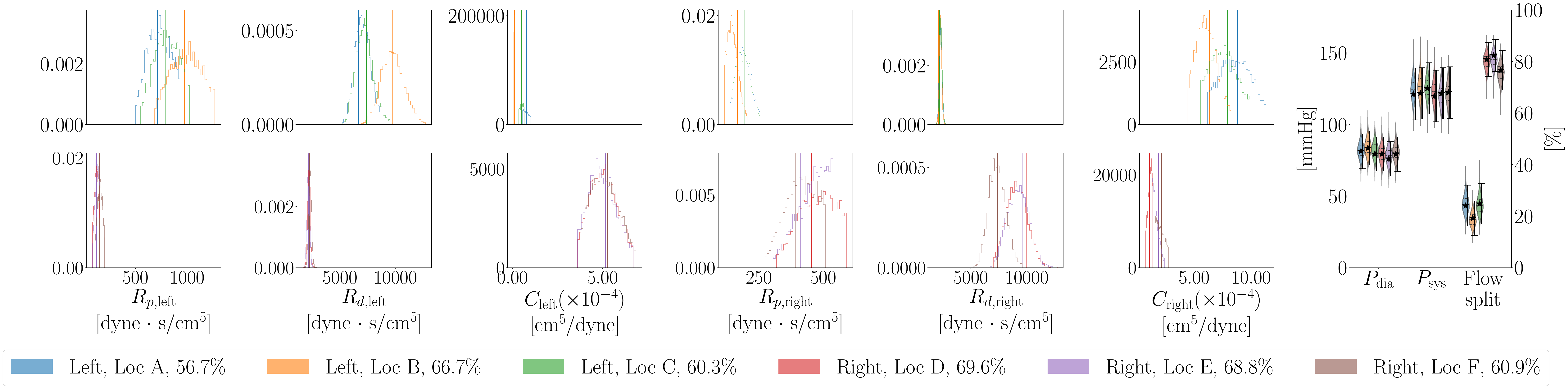}
    \caption{Marginal posteriors and corresponding true boundary conditions at one testing
sample. In the rightmost column, true observations are illustrated with black stars, with $\pm3$ standard deviations in black bars.}
    \label{fig:onehot_cfm_latent_estimation}
\end{figure}

Rather than using the six-dimensional latent space as a condition to the CFM model, we investigated whether the CFM model could \emph{infer} the latents. The tuned hyperparameters for the CFM model are found in ``One-hot with geometry estimation'' in \cref{tab:appendix_hyperparameters_MLP}. First, we show the results for the posterior distributions in \cref{fig:onehot_cfm_latent_estimation}, where we see that the overall distributions are comparable to those reported in \cref{sec:geometry_conditioning}. %Note that since explicit surface or mesh reconstruction is outside the scope of this work, the estimated boundary conditions were input to the ground truth zero-dimensional files to assess the accuracy (when compared with \cref{sec:geometry_conditioning}).

\begin{figure}[!htb]
    \centering
    \includegraphics[width=\linewidth]{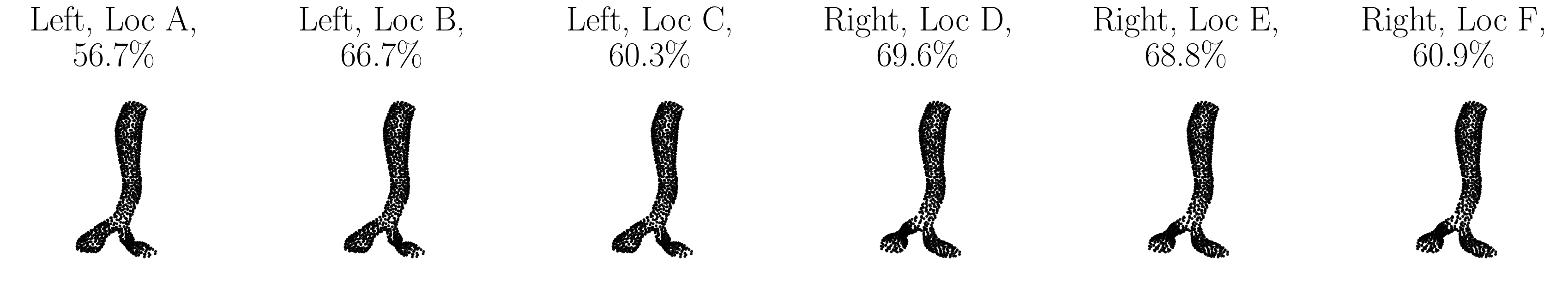}
    \caption{Reconstructed geometries from the estimated latents, which are then input to the decoder.}
    \label{fig:reconstructed_geoms_from_estimation_encoder}
\end{figure}

Next, we use the trained decoder from \cref{sec:geometry_conditioning} to decode the latents inferred by the CFM model. We plot the reconstructed geometries in \cref{fig:reconstructed_geoms_from_estimation_encoder}. We observe that the \emph{unsupervised} CFM model learns a six-dimensional latent representation that naturally separates into two approximately three-dimensional subspaces: one corresponding to the left iliac artery and one corresponding to the right iliac artery. This separation arises because the zero-dimensional simulations depend strongly on which iliac artery contains the stenosis. A stenosis in the left iliac artery alters the effective resistance of the blood vessel on the left, while a stenosis in the right iliac artery alters the resistance of the blood vessel on the right. These changes induce structurally distinct simulation outputs, since the lumped-parameter model computes blood vessel-averaged pressures and flow rates. As a result, geometries with left-sided stenosis and right-sided stenosis occupy different regions in output space. The CFM model, which is trained to match these simulation outputs, therefore learns to embed them in well-separated regions of latent space.

In contrast, the specific axial location of the stenosis along a given branch has a much weaker effect on the zero-dimensional simulation results. Since zero-dimensional models average quantities along each blood vessel and do not resolve spatial flow structure, variations in stenosis location along the same artery produce relatively similar outputs as long as the effective resistance is comparable. Consequently, the CFM model has little incentive to allocate separate latent directions to distinguish different axial positions. Instead, the learned latent representation primarily captures which artery is affected and the overall severity of the stenosis, while variations in location are weakly expressed.

% =========================================================
\section{Generation of inflow waveforms}\label{app:inflows}
% =========================================================

\begin{figure}[!htb]
    \centering
    \includegraphics[width=0.65\linewidth]{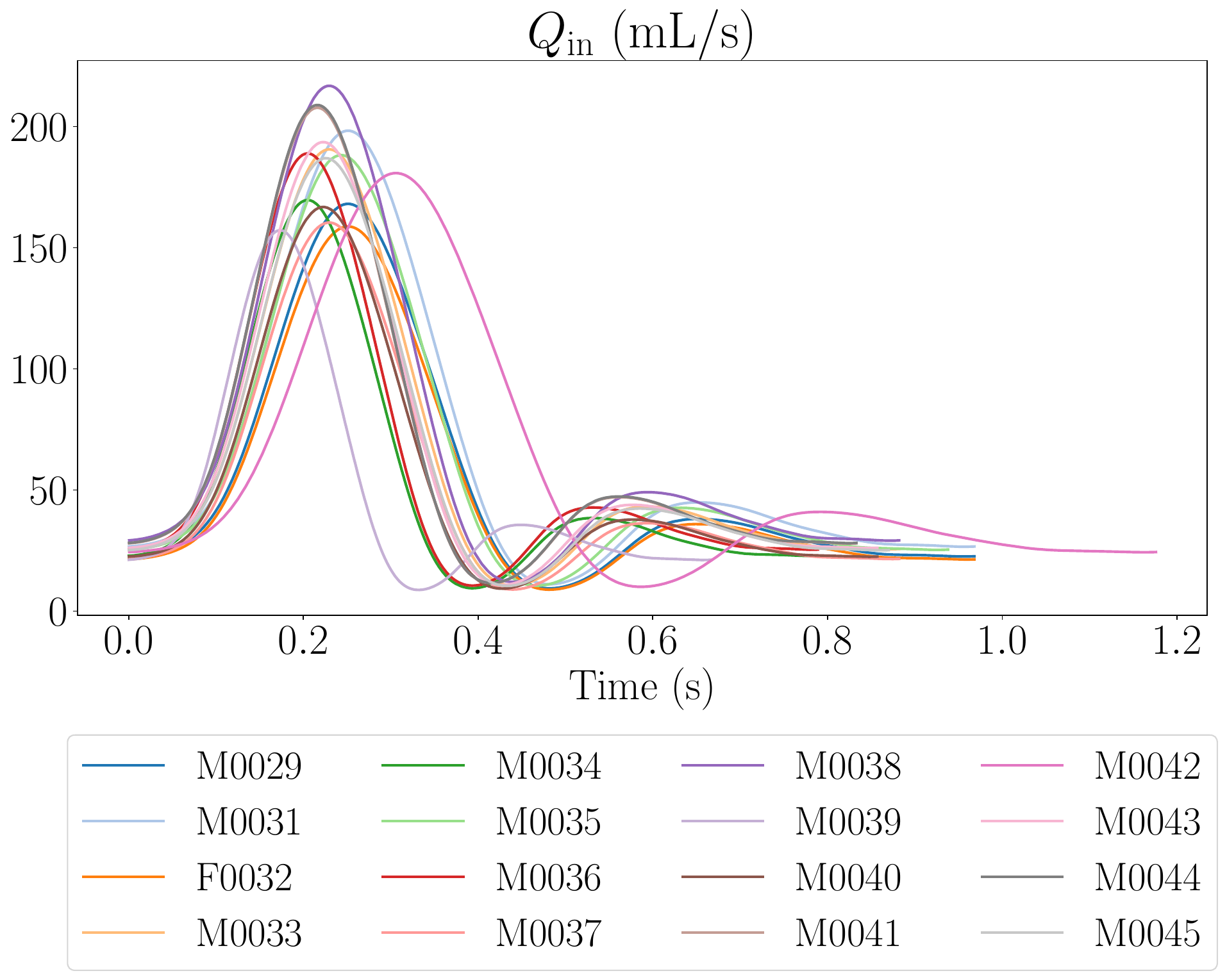}
    \caption{Flow profiles for different patients used for training data}
    \label{fig:flow_profiles_different_patients}
\end{figure}

We plot the inflow profiles for 16 abdominal aorta models available in the VMR in \cref{fig:flow_profiles_different_patients}. We split the data to 13 training samples and 3 testing samples.

\begin{figure}[!htb]
    \centering
    \includegraphics[width=\linewidth]{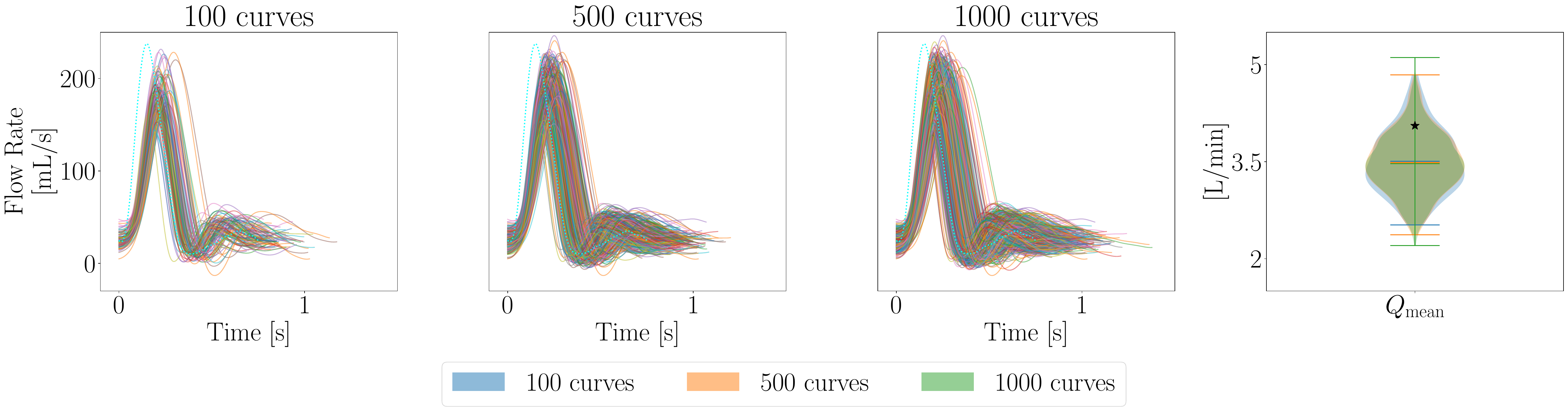}
    \caption{Inflow sampled from a flow matching algorithm, given the training dataset in \cref{fig:flow_profiles_different_patients}. These generated inflows are then used as training data for FalconBC.}
    \label{fig:generated_inflows}
\end{figure}

We train a flow matching algorithm separate from the FalconBC workflow to sample Fourier coefficients that are used to generate inflow curves. The sampled inflows are visualized in \cref{fig:generated_inflows}. Note that these were generated without a conditioning variable.

% ===============================================================
\section{3D vs 0D results using nominal values from optimization}
% ===============================================================

\begin{figure}[!htb]
    \centering
    \includegraphics[width=\linewidth]{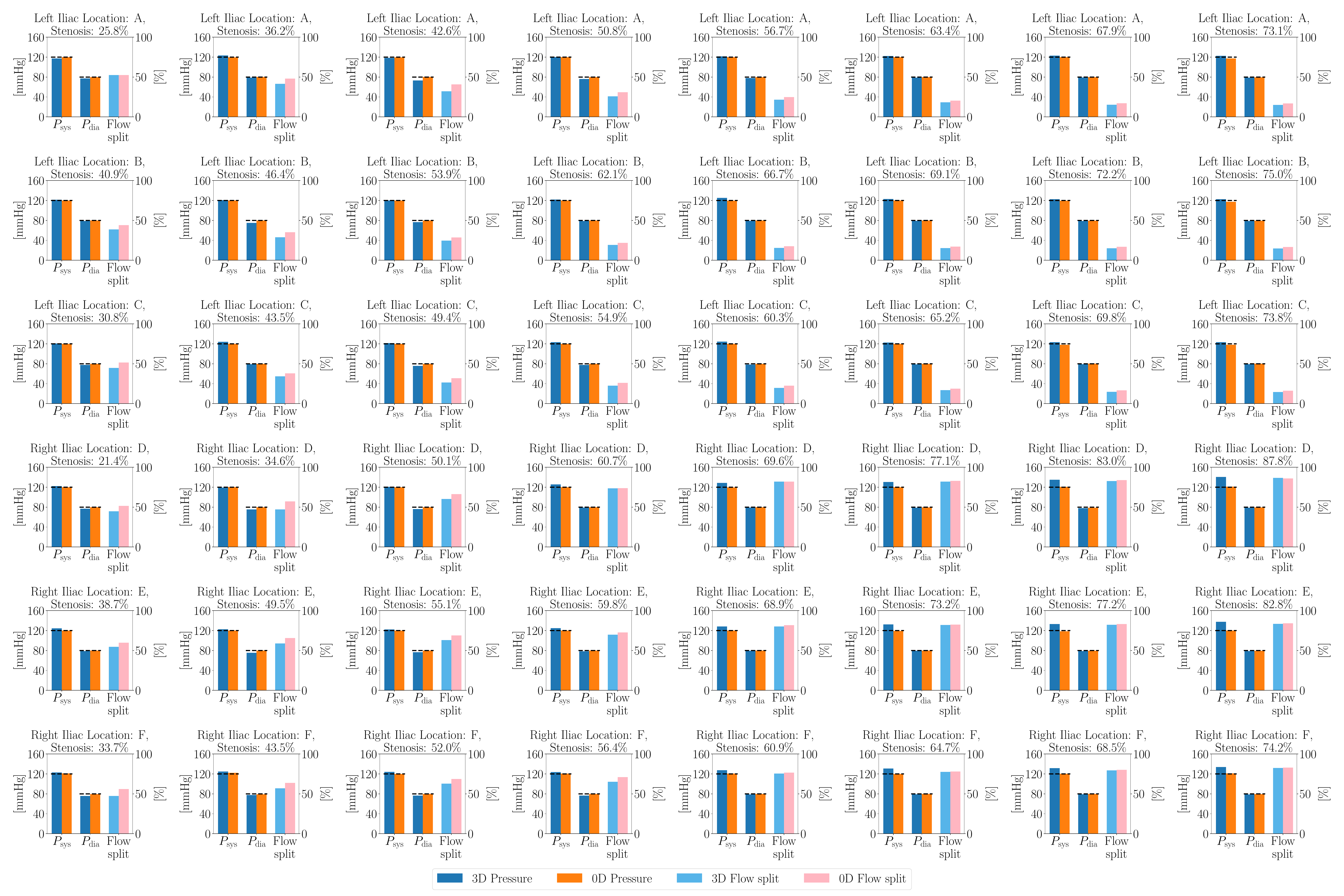}
    \caption{3D versus 0D results, where flow split indicates that going through the left iliac artery.}
    \label{fig:3d_vs_0d_results}
\end{figure}

To give the reader a sense of the close alignment between 3D and 0D results for the selected aorto-iliac bifurcation model, we run three-dimensional simulations using the nominal values obtained from optimization (details in \cref{app:geometry_bc_optimization}). 
We compare resulting systolic and diastolic pressure and flow split values in \cref{fig:3d_vs_0d_results}. Notice that the results are comparable with the exception of larger systolic pressure discrepancies for higher degrees of stenosis (in particular, those greater than 80\%).

\section{Details on geometry-aware optimization} \label{app:geometry_bc_optimization}

For each geometry, we solve the following constrained optimization problem.

The parameters to be optimized are:
\begin{equation}
\boldsymbol{x}
=
\big(
R_{p,\text{left}},\; R_{d,\text{left}},\; C_{\text{left}},\;
R_{p,\text{right}},\; R_{d,\text{right}},\; C_{\text{right}}
\big)
\end{equation}

\subsection{Optimization problem}

Given $\boldsymbol{x}$, the 0D solver produces (1) inlet pressure waveform $p(t;\boldsymbol{x})$ and (2) outlet flow waveforms $q_i(t;\boldsymbol{x})$ for each RCR boundary $i$.

We solve the following optimization problem:
\begin{equation}
\boldsymbol{x}^{*}
=
\arg\min_{\boldsymbol{x}}
\mathcal{L}(\boldsymbol{x}),
\end{equation}

where the losses to be minimized are pressure clinical targets, flow reversal, and monotonicity. 

\subsubsection{Optimization bounds}

The optimization is solved using differential evolution with geometry-dependent parameter bounds and warm-start continuation across increasing stenosis severity. For the first geometry in the continuation sequence, parameters are constrained using global physical bounds:
\begin{align}
R_{p,\min} \le &\, R_{p,\text{left}} \le R_{p,\max},
&
R_{d,\min} \le R_{d,\text{left}} \le R_{d,\max},
&
C_{\min} \le C_{\text{left}} \le C_{\max},
\\
R_{p,\min} \le &\, R_{p,\text{right}} \le R_{p,\max},
&
R_{d,\min} \le R_{d,\text{right}} \le R_{d,\max},
&
C_{\min} \le C_{\text{right}} \le C_{\max}.
\end{align}

For subsequent geometries, bounds are adaptively tightened using the solution
$\boldsymbol{\theta}^{\mathrm{prev}} =
(R_{p,\text{left}}^{\mathrm{prev}}, R_{d,\text{left}}^{\mathrm{prev}}, C_{\text{left}}^{\mathrm{prev}},
 R_{p,\text{right}}^{\mathrm{prev}}, R_{d,\text{right}}^{\mathrm{prev}}, C_{\text{right}}^{\mathrm{prev}})$
from the previous geometry.

Let $\alpha < 1$ and $\beta > 1$ denote the allowed fractional decrease and increase, respectively.
The continuation bounds are then defined as
\begin{align}
R_{p,\text{left}} &\in
\left[
R_{p,\text{left}}^{\mathrm{prev}},
\;
\min\!\left(R_{p,\max},\; \beta R_{p,\text{left}}^{\mathrm{prev}}\right)
\right],
\\
R_{d,\text{left}} &\in
\left[
R_{d,\text{left}}^{\mathrm{prev}},
\;
\min\!\left(R_{d,\max},\; \beta R_{d,\text{left}}^{\mathrm{prev}}\right)
\right],
\\
C_{\text{left}} &\in
\left[
C_{\min},
\;
\min\!\left(C_{\max},\; \beta C_{\text{left}}^{\mathrm{prev}}\right)
\right],
\\[6pt]
R_{p,\text{right}} &\in
\left[
\max\!\left(R_{p,\min},\; \alpha R_{p,\text{right}}^{\mathrm{prev}}\right),
\;
R_{p,\text{right}}^{\mathrm{prev}}
\right],
\\
R_{d,\text{right}} &\in
\left[
\max\!\left(R_{d,\min},\; \alpha R_{d,\text{right}}^{\mathrm{prev}}\right),
\;
R_{d,\text{right}}^{\mathrm{prev}}
\right],
\\
C_{\text{right}} &\in
\left[
\max\!\left(C_{\min},\; \alpha C_{\text{right}}^{\mathrm{prev}}\right),
\;
C_{\text{right}}^{\mathrm{prev}}
\right].
\end{align}

In all experiments, $\alpha = 0.9$ and $\beta = 1.1$, corresponding to a maximum
$10\%$ decrease or increase per continuation step.

\subsubsection{Optimization loss term}

The total loss minimized for each geometry is
\begin{equation}
\mathcal{L}(\boldsymbol{x})
=
\mathcal{L}_{\mathrm{pressure}}
+
10^{-4}\,\mathcal{L}_{\mathrm{flow}}
+
10^{-2}\,\mathcal{L}_{\mathrm{mono}}.
\end{equation}

We outline the three loss terms. The loss term for pressure is given by

\begin{equation}
\mathcal{L}_{\mathrm{pressure}}(\boldsymbol{x})
=
\left(
\frac{p_{\mathrm{sys}}(\boldsymbol{x}) - p_{\mathrm{sys}}^{*}}{p_{\mathrm{sys}}^{*}}
\right)^2
+
\left(
\frac{p_{\mathrm{dia}}(\boldsymbol{x}) - p_{\mathrm{dia}}^{*}}{p_{\mathrm{dia}}^{*}}
\right)^2,
\end{equation}

where 
$p_{\mathrm{sys}}(\boldsymbol{x}) = \max_t p (t)$ and $p_{\mathrm{dia}}(\boldsymbol{x}) = \min_t p (t)$ from the simulation, and the clinical targets are given by $p_{\mathrm{sys}}^{*}$ and $p_{\mathrm{dia}}^{*}$.

Next, for each outlet flow waveform $q_i(t)$, negative flow is penalized as
\begin{equation}
\mathcal{L}_{\mathrm{flow}}
=
\sum_i
\frac{1}{T}
\int_0^T
\bigl[\max(0,-q_i(t))\bigr]^2
\,dt .
\end{equation}

Finally, let $\bar{q}_{\text{left}}(\boldsymbol{x})$ denote the mean flow in the outlet of the stenosed branch (which could be either left or right),
\begin{equation}
\bar{q}_{\text{left}}(\boldsymbol{x})
=
\frac{1}{T}\int_0^T q_{\text{left}}(t)\,dt .
\end{equation}

Given the corresponding mean flow from the previous geometry $\bar{q}_{\text{left}}^{\mathrm{prev}}$, the monotonicity penalty term is
\begin{equation}
\mathcal{L}_{\mathrm{mono}}(\boldsymbol{x})
=
\max\!\left(0,\; \bar{q}_{\text{left}}(\boldsymbol{x}) - \bar{q}_{\text{left}}^{\mathrm{prev}}\right)^2 .
\end{equation}

\subsection{Physical constraints}

Additionally, we apply hard physical constraints:

\begin{align}
C_{\text{left}},C_{\text{right}} &\in [C_{\text{min}},C_{\text{max}}] \\[4pt]
\tau_{\text{left}} &= R_{d,\text{left}}\,C_{\text{left}} \in [\tau_{\text{min}}, \tau_{\text{max}}],
\\
\tau_{\text{right}} &= R_{d,\text{right}}\,C_{\text{right}} \in [\tau_{\text{min}}, \tau_{\text{max}}],
\\[4pt]
\end{align}

where $C_{\min} ,  C_{\max} \in [10^{-5},10^{-2}]$, $\tau_{\text{min}} = 0.05$, and $\tau_{\text{max}} = 10$. Violation of any constraint results in a large penalty value.

\end{document}